\pdfoutput=1

\documentclass[11pt]{article}

\usepackage[final]{acl}

\usepackage{times}
\usepackage{latexsym}

\usepackage[T1]{fontenc}

\usepackage[utf8]{inputenc}

\usepackage{microtype}

\usepackage{inconsolata}

\usepackage{graphicx}

\usepackage{bbm} 
\usepackage{amsfonts}
\usepackage{enumitem}
\usepackage{makecell}
\usepackage{multirow}
\usepackage{adjustbox}

\usepackage[edges]{forest}
\usepackage{amsmath}
\usepackage{tikz}
\usepackage{multicol}
\usetikzlibrary{arrows,shapes,positioning,shadows,trees}

\usepackage{tabularx}
\usepackage{threeparttable}
\usepackage[nointegrals]{wasysym}
\usepackage{booktabs}
\usepackage{array}

\usepackage{ragged2e}
\newcolumntype{Y}{>{\arraybackslash}X}
\newcolumntype{Z}{>{\centering\arraybackslash}p{1cm}}
\newcommand*\rot[1]{\hbox to1em{\hss\rotatebox[origin=br]{-12}{#1}}}

\usepackage[most]{tcolorbox}
\usepackage{xcolor}

\newtcolorbox{scopebox}{colback=gray!10!white,
  colframe=black!50!white,
  boxrule=0.4pt, 
  left=4pt,right=4pt, 
  top=4pt,bottom=4pt,
  before skip=8pt, 
  after skip=8pt,
} 

\newtcolorbox{insightcallout}[2][]{%
  enhanced,
  breakable,
  colback=gray!10!white,
  boxrule=0pt,
  frame hidden,
  borderline west={3pt}{0pt}{myblue},
  left=6pt,right=3pt, 
  top=4pt,bottom=4pt,
  before skip=8pt, 
  after skip=8pt,
  sharp corners,
  title={#2},
  coltitle=black,
  attach title to upper,
  #1
}
\definecolor{mypink}{HTML}{FFE4E5}
\definecolor{mypurple}{HTML}{E5D0E9}
\definecolor{mygreen}{HTML}{BBDED6}
\definecolor{myyellow}{HTML}{FFE083}
\definecolor{myblue}{HTML}{1E88E5}

\definecolor{mygreen-deep1}{HTML}{339C84}
\definecolor{mygreen-deep2}{HTML}{77BDAD}
\definecolor{mypurple-deep1}{HTML}{B172BD}
\definecolor{mypurple-deep2}{HTML}{CBA1D3}
\definecolor{mypink-deep1}{HTML}{FFAEB1}
\definecolor{mypink-deep2}{HTML}{FFC9CB}

\usepackage{fontawesome} 
\usepackage{makecell,colortbl}
\newcommand{\Strong}{\textcolor{myblue}{\CIRCLE}}        
\newcommand{\Weak}{\textcolor{myblue}{\Circle}}          
\newcommand{\primary}{\textcolor{mygreen-deep2}{\CIRCLE}}
\newcommand{\secondary}{\textcolor{mypurple-deep2}{\LEFTcircle}}      
\newcommand{\istar}{\textcolor{myblue}{$\star$}}

\definecolor{dense4}{RGB}{255,250,220} 
\definecolor{dense3}{RGB}{255,236,158} 
\definecolor{dense2}{RGB}{255,213,79}  
\definecolor{dense1}{RGB}{255,179,0} 
\newcommand{\heat}[1]{\cellcolor{#1}\makebox[1.2em]{}}

\newcommand{\RowDensityLegend}{
\begin{tikzpicture}[baseline]
  \def\w{0.9} \def\h{0.18}
  \node[anchor=east] at (-0.25,0.5*\h) {\scriptsize\textit{Row density}};
  \foreach \i/\col in {0/dense4,1/dense3,2/dense2,3/dense1}{
    \fill[\col] (\i*\w,0) rectangle ++(\w,\h);
    \draw[black!20] (\i*\w,0) rectangle ++(\w,\h);
  }
  \foreach \i/\lab in {0/{0--10},1/{11--20},2/{21--30},3/{31+}}{
    \node[anchor=north] at (\i*\w+0.5*\w, -0.02) {\scriptsize \lab};
  }
\end{tikzpicture}
}

\newcommand{\ColDensityLegend}{
\begin{tikzpicture}[baseline]
  \def\w{0.9} \def\h{0.18}
  \node[anchor=east] at (-0.25,0.5*\h) {\scriptsize\textit{Column density}};
  \foreach \i/\col in {0/dense4,1/dense3,2/dense2,3/dense1}{
    \fill[\col] (\i*\w,0) rectangle ++(\w,\h);
    \draw[black!20] (\i*\w,0) rectangle ++(\w,\h);
  }
  \foreach \i/\lab in {0/{0--20},1/{21--40},2/{41--60},3/{61+}}{
    \node[anchor=north] at (\i*\w+0.5*\w, -0.02) {\scriptsize \lab};
  }
\end{tikzpicture}
}

\newcommand{\TextLegend}{
  \raisebox{0.75\height}{%
    \begin{minipage}[t]{11em}
      \scriptsize
      \textit{Behavior dimension:}\\
      \colorbox{mypink!60}{temporal}, \colorbox{mypurple!60}{spatial}, \colorbox{mygreen!60}{structural}.
    \end{minipage}
  }
}

\usepackage{pifont}
\newcommand{\cmark}{\ding{51}}

\definecolor{tagpurple}{HTML}{6A1B9A}
\definecolor{taggrey}{HTML}{616161} 

\newcommand{\pill}[2]{%
  \tikz[baseline=(X.base)]\node[
    rounded corners=1.6pt,
    fill=#1,
    text=white,
    font=\scriptsize\bfseries,
    inner xsep=0.55em,
    inner ysep=0.18em
  ] (X) {#2};%
}

\newcommand{\tagA}{\pill{dense1}{A}}
\newcommand{\tagS}{\pill{myblue}{S}}
\newcommand{\tagQ}{\pill{tagpurple}{Q}}
\newcommand{\tagF}{\pill{dense1}{F}}
\newcommand{\tagB}{\pill{tagpurple}{B}}
\newcommand{\tagR}{\pill{taggrey}{R}}
\newcommand{\tagRR}{\pill{dense1}{R}}

\newcommand{\rev}[1]{\textcolor{black}{#1}}
\newcommand{\revm}[1]{\textcolor{black}{#1}}

\title{Towards Efficient Large Language Model Serving:\\A Survey on System-Aware KV Cache Optimization}

\author{
 \textbf{Jiantong Jiang\textsuperscript{1}},
 \textbf{Peiyu Yang\textsuperscript{1}}\thanks{Peiyu Yang is the corresponding author.},
 \textbf{Rui Zhang\textsuperscript{2}},
 \textbf{Feng Liu\textsuperscript{1}}
\\
 \textsuperscript{1}School of Computing and Information Systems, The University of Melbourne, \\
 \textsuperscript{2}School of Computer Science and Technology, \\ Huazhong University of Science and Technology \footnotesize (www.ruizhang.info)
\\
 \texttt{\{jiantong.jiang, peiyu.yang\}@unimelb.edu.au} \\ \texttt{rayteam@yeah.net, fengliu.ml@gmail.com}
\\ \href{https://github.com/jjiantong/Awesome-KV-Cache-Optimization}{
  \includegraphics[height=1em]{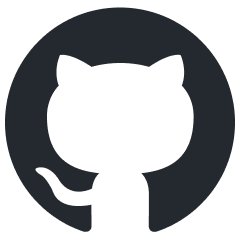}\; 
  \textcolor{magenta}{\texttt{https://github.com/jjiantong/Awesome-KV-Cache-Optimization}}
}
}

\begin{document}
\maketitle
\begin{abstract}
Despite the rapid advancements of large language models (LLMs), LLM serving systems remain memory-intensive and costly. The key-value (KV) cache, which stores KV tensors during autoregressive decoding, is crucial for enabling low-latency, high-throughput LLM inference serving. 
In this survey, we focus on \underline{s}ystem-aware \underline{K}V \underline{i}nfrastructure for \underline{s}erving LLMs (abbreviated as \textit{sKis}). We revisit recent work from a system behavior perspective, organizing existing efforts into three dimensions: execution and scheduling (temporal), placement and migration (spatial), and representation and retention (structural). Furthermore, we analyze \rev{cross-behavior co-design affinity and behavior-objective links}, highlighting future opportunities. Our work systematizes a rapidly evolving area, providing a foundation for understanding and innovating KV cache designs in modern LLM serving infrastructure. 
\end{abstract}


\section{Introduction}
\label{sec_intro}

Large language models (LLMs) have showcased exceptional abilities across diverse applications~\cite{zhao2023survey}, with notable examples like GPT~\cite{radford2018improving,radford2019language,brown2020language,achiam2023gpt}, LLaMA~\cite{touvron2023llama,touvron2023llama2}, and OPT~\cite{zhang2022opt}. These models excel at large-scale high-quality language understanding and generation, and their extensions also support plenty of multi-modal generation tasks~\cite{zhang2025comprehensive, zhang2025overview}.
These abilities are largely powered by the Transformer architecture~\cite{vaswani2017attention}, which efficiently captures long-range dependencies via self-attention.


Despite their success, serving LLMs efficiently remains non-trivial~\cite{li2024llm}. Transformer-based LLMs generate tokens autoregressively, with each token conditioned on all previous ones. 
To avoid redundant compute, serving systems adopt a \textit{key-value (KV) cache}~\cite{pope2023efficiently} to store intermediate KV tensors of the generated tokens.
Yet, as prompt and output length grow, the KV cache can reach millions of tokens~\cite{longrope}, creating memory bottlenecks and highlighting the critical role of KV cache optimization.
Thus, a growing body of KV-centric techniques has emerged, yielding memory savings and efficiency gains in throughput and latency~\cite{li2024survey}.

\begin{figure}
    \centering
    \includegraphics[width=1.0\linewidth]{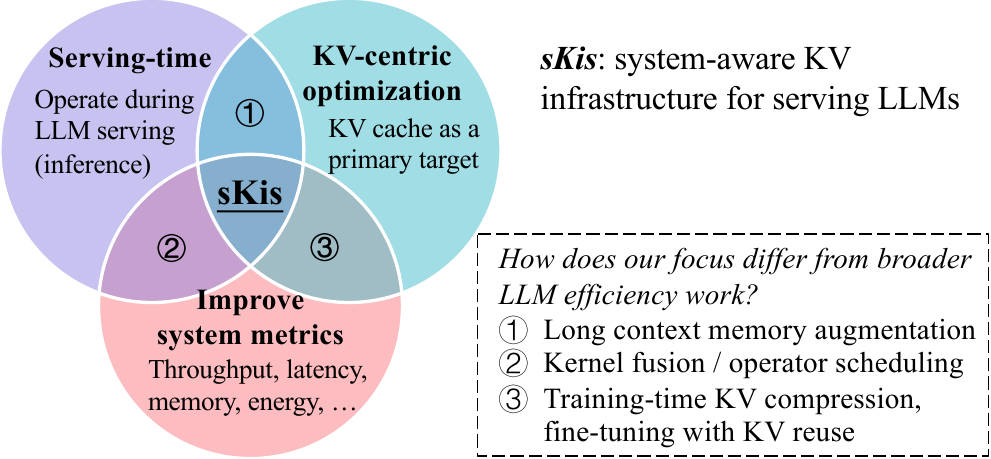}
    \caption{Positioning of the survey scope (``sKis'').}
    \label{fig_pos}
\end{figure}

\begin{figure}
    \centering
    \includegraphics[width=1.0\linewidth]{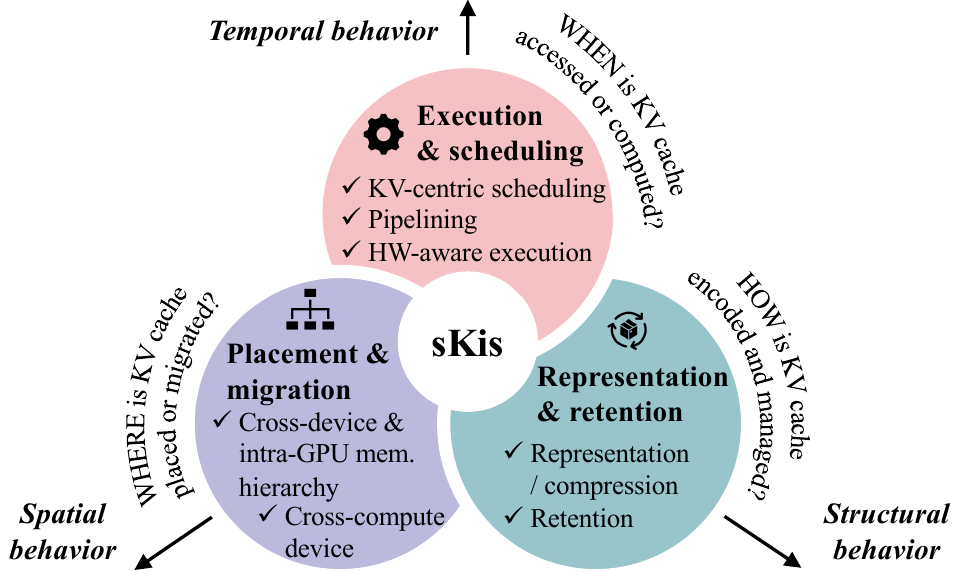}
    \vspace{-4mm}
    \caption{Taxonomy of the survey that covers temporal, spatial, and structural dimensions.
    }
    \label{fig_3_d}
\end{figure}

To this end, we argue that it deserves a deep investigation of system-aware, serving-time, KV-centric optimization methods, as shown in Fig.~\ref{fig_pos},
which we call this scope \textit{sKis}.
We adopt a system-oriented taxonomy to offer a comprehensive understanding of sKis, categorizing methods along three fundamental axes of system behaviors, as shown in Fig.~\ref{fig_3_d}:
(i) \textbf{execution and scheduling} focuses on the \textit{temporal} control of when KV data is accessed, computed, or scheduled (cf. \S~\ref{sec_comp}); (ii) \textbf{placement and migration} captures the \textit{spatial} decisions of where KV data is placed or moved across memory tiers or devices (cf. \S~\ref{sec_move}); and (iii) \textbf{representation and retention} concerns the \textit{structural} treatment of how KV data is compressed or managed (cf. \S~\ref{sec_mem}).
%
%
We further analyze cross-behavior \rev{co-design patterns} and behavior–objective effects to reveal overlooked regions and open challenges (cf. \S~\ref{sec_observation_and_challenge}).

\begin{table}
\centering
\footnotesize
\setlength{\tabcolsep}{2pt}
\caption{\revm{Comparison of our work with surveys related to efficient LLM inference or serving.}}
\vspace{-2mm}
\label{tab_related_work}
\resizebox{\columnwidth}{!}{
\revm{
\begin{tabular}{lcccl}
\toprule
Survey & \makecell{KV-\\centric} & \makecell{Serving\\only} & \makecell{No\\retrain} & \makecell{Organizing\\principle} \\
\midrule
\citet{miao2023towards} &   & \cmark &  & Algorithm-system\\
\citet{yuan2024llm} &   & \cmark &  & Optimization layer \\
\citet{li2024llm} &   & \cmark & \cmark & System component \\
\citet{zhou2024survey} &   & \cmark &  & Optimization layer  \\
\citet{zhen2025taming} &   & \cmark & \cmark & Serving scale \\
\citet{shi2024keep} & \cmark &  &  & Lifecycle stage \\
\citet{li2024survey} & \cmark & \cmark &  & Optimization layer\\
\citet{liu2025kv} & \cmark & \cmark &  & Compression types \\
\midrule
This survey (sKis) & \cmark & \cmark &  & System behavior \\
\bottomrule
\end{tabular}
}
}
\end{table}

\revm{While prior surveys span efficient LLM inference and serving~\cite{zhou2024survey, yuan2024llm, miao2023towards, li2024llm, zhen2025taming}, they are general surveys where the KV cache is discussed only as a minor component. KV-specific surveys are closest to our topic~\cite{shi2024keep, li2024survey, liu2025kv}, but they typically organize by lifecycle stages or optimization layers.}
Instead, this survey focuses exclusively on sKis and distinguishes itself by offering a novel behavior-oriented perspective and a deeper understanding. \revm{We compare related surveys in Tab.~\ref{tab_related_work} and provide further details in App.~\ref{a_related_work}.}

To the best of our knowledge, we are the first to frame KV cache optimization as a temporal-spatial-structural behavior space, enabling principled analysis and actionable future directions. \rev{Because this design space is decoupled from model and kernel details, it also offers a stable lens for situating new techniques in this rapidly evolving area. }



\section{Foundations, Scope and Taxonomy}
\label{sec_pre}





\begin{figure}[t]
    \centering
    \includegraphics[width=1.0\linewidth]{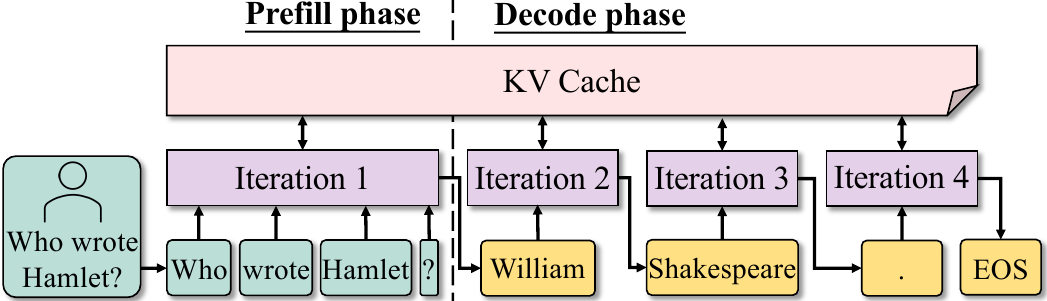}
    \vspace{-4mm}
    \caption{Prefill and decode phases of LLM inference.}
    \label{fig_llm_inference}
\end{figure}


\paragraph{LLM Inference and KV Cache.}
LLMs generate tokens autoregressively, as shown in Fig.~\ref{fig_llm_inference} (\revm{see preliminaries on Transformer-based LLMs in App.~\ref{a_llm}}). At each step, the model consumes the input and previously generated tokens to generate the next token. This process has two phases: \textit{prefill} processes the initial input and generates the first output token, and \textit{decode} generates tokens autoregressively.
Due to the quadratic cost of self-attention, repeatedly computing attention across tokens is expensive. To this end, \textit{key-value (KV) cache} is used to store the intermediate KV tensors computed previously, allowing the model to efficiently reuse them without recomputing attention over the entire sequence. 





\paragraph{Scope.}
This survey investigates recent advances in the sKis scope, as shown in Fig.~\ref{fig_pos}. 
%

\begin{scopebox} 
\textit{sKis denotes \underline{s}ystem-aware \underline{K}V \underline{i}nfrastructure for \underline{s}erving LLMs. A method belongs to sKis if it: (i) operates during serving (inference), (ii) centers on KV caches as the primary optimization target, and (iii) aims to improve system metrics without retraining the base LLM's weights or modifying its Transformer architecture.}
\end{scopebox}


We acknowledge related but out-of-scope directions that redesign the attention or KV mechanism~\citep{deepseekv2, yuan2025native, sun2026bat}, or modify the backbone model structure~\citep{ashkboos2024slicegpt, li2024sglp, lu2026reassessing}. Although these directions can reshape the KV footprint and improve efficiency, they do so by altering the model itself or requiring model adaptation. In contrast, this survey focuses on serving-time KV cache optimizations for off-the-shelf models, which remains a common deployment setting in the community.

\paragraph{Taxonomy.}
We organize literature on sKis by low-level system behaviors, as shown in Fig.~\ref{fig_3_d}. Further details are provided in App.~\ref{a_taxonomy}.
%
However, similar to how a modern OS includes components for scheduling, memory, and I/O, LLM serving systems often involve techniques spanning various system dimensions. Thus, a single paper may naturally touch on several categories. 
For clarity and focus, we mention 1-2 primary categories per work based on its main contributions. Minor associations are not elaborated, and we refer to App.~\ref{a_extended_c} for details. 
We summarize the organization and methods in Fig.~\ref{fig_taxonomy} and the key takeaways in App.~\ref{a_takeaway}.

\def\HierarchyList{FlexGen \cite{flexgen}, FastServe \cite{fastserve}, 
ALISA \cite{alisa}, InfiniGen \cite{infinigen}, CachedAttention \cite{cachedattention}, D{\'e}j{\`a}Vu \cite{dejavu}, LayerKV \cite{layerkv}, FastSwitch \cite{fastswitch}, InfLLM \cite{infllm}, ArkVale \cite{arkvale}, IMPRESS \cite{impress}, Pensieve \cite{pensieve}, OmniKV \cite{omnikv}, ClusterKV \cite{clusterkv}, PQCache \cite{pqcache}, ShadowKV \cite{shadowkv}, SpeCache \cite{specache}, SlimInfer \cite{sliminfer}, RAGCache \cite{ragcache}, LMCache \cite{lmcache}, RetrievalAttention \cite{retrievalattention}, KVFlow \cite{kvflow}, TraCT \cite{tract}, CXL-SpecKV \cite{cxl-speckv}, Beluga \cite{beluga}}

\def\EvictList{H$_2$O \cite{h2o}, Scissorhands \cite{scissorhands}, RoCo \cite{roco}, FastGen \cite{fastgen}, StreamingLLM \cite{streamingllm}, Keyformer \cite{keyformer}, PyramidKV \cite{pyramidkv}, NACL \cite{nacl}, PyramidInfer \cite{pyramidinfer}, BUZZ \cite{buzz}, TOVA \cite{tova}, VATP \cite{vatp}, L2KV \cite{l2kv}, SnapKV \cite{snapkv}, CAKE \cite{cake}, D$_2$O \cite{d2o}, SepLLM \cite{sepllm}, LaCache \cite{lacache}, KVCompose \cite{kvcompose}, DiffKV \cite{diffkv}, EvolKV \cite{evolkv}, DynamicKV \cite{dynamickv}, Ada-KV \cite{adakv}}

\def\QuantList{SmoothQuant \cite{smoothquant}, FlexGen \cite{flexgen}, WKVQuant \cite{wkvquant}, MiKV \cite{mikv}, QAQ \cite{qaq}, Atom \cite{atom}, KIVI \cite{kivi}, CacheGen \cite{cachegen}, DecoQuant \cite{decoquant}, GEAR \cite{gear}, SKVQ \cite{skvq}, KVQuant \cite{kvquant}, CQ \cite{cq}, ZipCache \cite{zipcache}, QJL \cite{qjl}, VQ-LLM \cite{vqllm}, SQuat \cite{squat}, QoQ \cite{qserve}, CommVQ \cite{commvq}, OTT \cite{ott}, NSNQuant \cite{nsnquant}, ChanMix \cite{chanmix}}

\tikzstyle{my-box}=[
    rectangle,
    rounded corners,
]
\tikzstyle{leaf1}=[my-box, 
    fill=mypink!30, align=left, text width=42.4em,
]
\tikzstyle{leaf2}=[my-box, 
    fill=mypurple!30, align=left, 
]
\tikzstyle{leaf3}=[my-box, 
    fill=mygreen!30, align=left, text width=37.9em, 
]

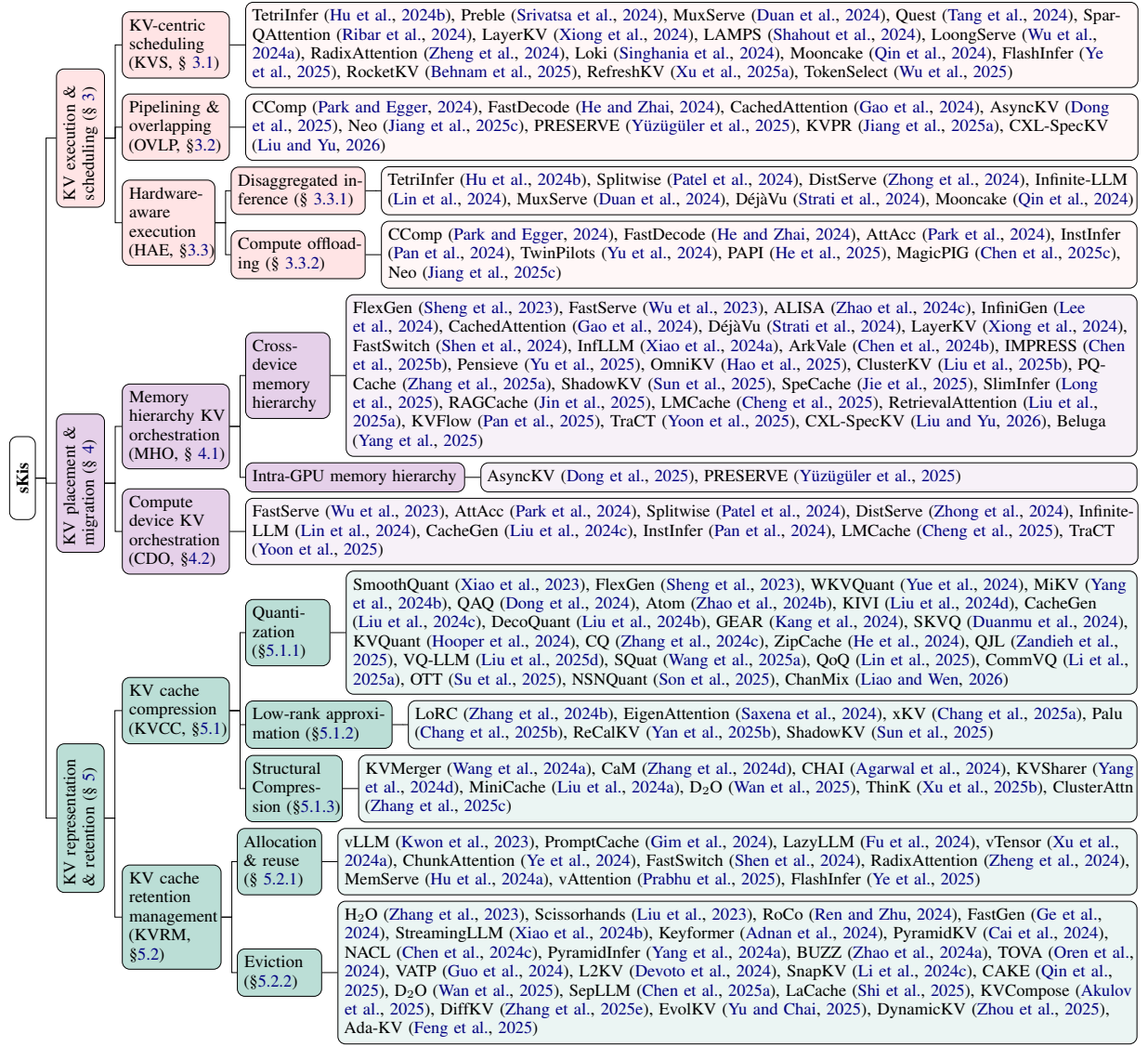
\begin{figure*}[t!]
    \centering
    \resizebox{1.01\textwidth}{!}{%
    \begin{forest}
        forked edges,
        for tree={
            grow=east,
            reversed=true,
            anchor=base west,
            font=\small,
            rectangle,
            draw=black,
            rounded corners,
            minimum width=4em,
            s sep=3pt,
            l sep=0.75em,
            ver/.style={rotate=90, anchor=center},
        },
        where level=1{text width=6.0em,}{},
        where level=2{text width=4.5em, xshift=-0.8em}{}, 
        where level=3{text width=3.4em,}{},
        [
            \textbf{sKis}, ver  
            [
                KV execution \&\\ scheduling (\S~\ref{sec_comp}), ver, fill=mypink,
                [
                    \parbox{5.3em}{KV-centric scheduling (KVS, \S~\ref{sec_scheduling})}, fill=mypink, text width=4.5em,
                    [
                        \parbox{51.4em}{TetriInfer \cite{tetriinfer}, Preble \cite{preble}, MuxServe \cite{muxserve}, Quest \cite{quest}, SparQAttention \cite{sparq}, LayerKV \cite{layerkv}, LAMPS \cite{lamps},  LoongServe \cite{loongserve},  RadixAttention \cite{radixattention}, Loki \cite{loki}, Mooncake \cite{mooncake}, FlashInfer \cite{flashinfer}, RocketKV \cite{rocketkv}, RefreshKV \cite{refreshkv}, TokenSelect \cite{tokenselect}}, leaf1
                    ]
                ]                
                [
                    \parbox{5.5em}{Pipelining \& overlapping (OVLP, \S \ref{sec_po})}, fill=mypink
                    [
                        \parbox{51.4em}{CComp \cite{cpucompv2}, FastDecode \cite{fastdecode}, CachedAttention \cite{cachedattention}, AsyncKV \cite{asynckv}, Neo \cite{neo}, PRESERVE \cite{preserve}, KVPR \cite{kvpr}, CXL-SpecKV \cite{cxl-speckv}}, leaf1
                    ]
                ]
                [
                    \parbox{5.1em}{Hardware-aware execution (HAE, \S \ref{sec_hs})}, fill=mypink, text width=3.8em
                    [
                        \parbox{7.2em}{Disaggregated inference (\S~\ref{sec_di})}, fill=mypink, text width=5.8em
                        [
                            \parbox{44.5em}{TetriInfer \cite{tetriinfer}, Splitwise \cite{splitwise}, DistServe \cite{distserve}, Infinite-LLM \cite{infinitellm},  MuxServe \cite{muxserve}, D{\'e}j{\`a}Vu \cite{dejavu}, Mooncake \cite{mooncake}}, leaf1, text width=35.8em,
                        ]
                    ]
                    [
                        \parbox{7.5em}{Compute offloading (\S~\ref{sec_comp_offload})}, fill=mypink, text width=5.8em
                        [
                            \parbox{44.0em}{CComp \cite{cpucompv2}, FastDecode \cite{fastdecode}, AttAcc \cite{attacc}, InstInfer \cite{instinfer}, TwinPilots \cite{twinpilots}, PAPI \cite{papi}, MagicPIG \cite{magicpig}, Neo \cite{neo}}, leaf1, text width=35.8em,
                        ]
                    ]                  
                ]
            ]
            [
                KV placement \&\\migration (\S~\ref{sec_move}), ver, fill=mypurple,
                [
                    \parbox{5.7em}{Memory hierarchy KV orchestration (MHO, \S~\ref{sec_hierarchy})}, fill=mypurple, 
                    [
                        \parbox{3.8em}{Cross-device memory hierarchy}, fill=mypurple, 
                        [
                            \parbox{45.4em}{\HierarchyList}, leaf2, text width=37.5em,
                        ]
                    ]
                    [
                        \parbox{16.5em}{Intra-GPU memory hierarchy}, fill=mypurple, text width=9.9em
                        [
                            \parbox{38.5em}{AsyncKV \cite{asynckv}, PRESERVE \cite{preserve}}, leaf2, text width=31.0em,
                        ]
                    ]
                ]
                [
                    \parbox{5.7em}{Compute device KV orchestration (CDO, \S \ref{sec_m})}, fill=mypurple, 
                    [
                        \parbox{51.6em}{FastServe \cite{fastserve}, AttAcc \cite{attacc}, Splitwise \cite{splitwise}, DistServe \cite{distserve}, Infinite-LLM \cite{infinitellm}, CacheGen \cite{cachegen}, InstInfer \cite{instinfer}, LMCache \cite{lmcache}, TraCT \cite{tract}}, leaf2, text width=42.4em
                    ]
                ]
            ]
            [
                KV representation\\\& retention (\S~\ref{sec_mem}), ver, fill=mygreen, text width=6.5em
                [
                    \parbox{6.3em}{KV cache compression (KVCC, \S \ref{sec_kvcc})}, fill=mygreen,
                    [
                        \parbox{5em}{Quanti-zation (\S \ref{sec_kvq})}, fill=mygreen, 
                        [
                            \parbox{45.4em}{\QuantList}, leaf3, text width=37.5em, 
                        ]
                    ]
                    [
                        \parbox{8.2em}{Low-rank approximation (\S \ref{sec_kvla})}, fill=mygreen, text width=6.4em
                        [
                            \parbox{43.4em}{LoRC \cite{lorc}, EigenAttention \cite{eigenattention}, xKV \cite{xkv}, Palu \cite{palu}, ReCalKV \cite{recalkv}, ShadowKV \cite{shadowkv}}, leaf3, text width=34.5em,
                        ]
                    ]
                    [
                        \parbox{5.2em}{Structural Compression (\S \ref{sec_kvsc})}, fill=mygreen, text width=4.0em
                        [
                            \parbox{45.1em}{KVMerger \cite{kvmerger}, CaM \cite{cam}, CHAI \cite{chai}, KVSharer \cite{kvsharer}, MiniCache \cite{minicache}, D$_2$O \cite{d2o}, ThinK \cite{think}, ClusterAttn \cite{clusterattn}}, leaf3, text width=36.9em,
                        ]
                    ]
                ]
                [
                    \parbox{5.2em}{KV cache retention management (KVRM, \S \ref{sec_clm})}, fill=mygreen, text width=4.1em 
                    [
                        \parbox{5em}{Allocation \& reuse (\S~\ref{sec_kvar})}, fill=mygreen, 
                        [
                            \parbox{46.2em}{vLLM~\cite{vllm}, PromptCache \cite{promptcache}, LazyLLM \cite{lazyllm}, vTensor \cite{vtensor}, ChunkAttention \cite{chunkattention}, FastSwitch \cite{fastswitch}, RadixAttention \cite{radixattention}, MemServe \cite{memserve}, vAttention \cite{vattention},  FlashInfer \cite{flashinfer}}, leaf3, 
                        ]
                    ]
                    [
                        \parbox{5em}{Eviction (\S \ref{sec_kve})}, fill=mygreen, 
                        [
                            \parbox{46.2em}{\EvictList}, leaf3, 
                        ]
                    ]
                ]
            ]  
        ]
    \end{forest}
    }
    \vspace{-6mm}
    \caption{Taxonomy of sKis and associated methods. Each method is annotated with its primary contributions for conciseness. Minor category associations are omitted here and listed in App.~\ref{a_extended_c}, Tab.~\ref{tab_a_extended_c}.}
    \label{fig_taxonomy}
    \vspace{-0.5mm}
\end{figure*}

\section{KV Execution and Scheduling}
\label{sec_comp}

This section captures the temporal behaviors of KV cache usage, including how cache entries are scheduled and executed efficiently at runtime.

\subsection{KV-centric Scheduling}
\label{sec_scheduling}

While scheduling is a long-studied system problem, KV-centric scheduling (\textit{KVS}) methods explicitly integrate KV characteristics into runtime decisions.

At the \textbf{request level}, some methods adopt KV usage-aware scheduling to balance resource load and reduce contention~\cite{tetriinfer, muxserve, layerkv, lamps, loongserve}. For example, TetriInfer~\cite{tetriinfer} prioritizes requests using predicted KV usage to mitigate prefill-decode interference.
Another line is reuse-aware, prioritizing high-reuse requests to maximize KV cache hit rate~\cite{radixattention} or using KV reuse potential as a key signal in decisions~\cite{preble, mooncake}.


At finer granularity, \textbf{token-level} methods decide which KV entries participate in attention based on estimated contributions~\cite{quest, sparq, loki, rocketkv, refreshkv, tokenselect}, for example via periodic refresh that alternates full-context and subset attention~\cite{refreshkv}.
%
At the \textbf{kernel level}, methods like FlashInfer~\cite{flashinfer} schedule attention workloads across CUDA thread blocks based on query and KV lengths. 

\begin{table*}[t]
\centering
\footnotesize
\setlength{\tabcolsep}{4pt}
\caption{Summary of OVLP methods. \revm{``Comp'' denotes compute, ``I/O'' denotes KV data movement (host–device transfer or on-device memory movement), and ``comm'' denotes collective communication.}}
\label{tab_overlap}
\resizebox{\textwidth}{!}{
\begin{tabular}{lllc}
\toprule
Mode & Method & Overlapped operations (with transfer path) & Granularity \\
\midrule
\multirow{2}{*}{Comp--Comp} & FastDecode~\cite{fastdecode} & CPU R-part comp $\leftrightarrow$ GPU S-part comp & Token-wise \\
& Neo~\cite{neo} & CPU attention comp $\leftrightarrow$ GPU linear ops & Sub-batch-wise \\
\midrule
\multirow{4}{*}{Comp--I/O} & CComp~\cite{cpucompv2} & CPU MHSA comp $\leftrightarrow$ FFN data transfer (CPU$\rightarrow$GPU) & Split point \\
& CachedAttention~\cite{cachedattention} & GPU comp $\leftrightarrow$ KV load/store (CPU$\leftrightarrow$GPU) & Layer-wise \\
& AsyncKV~\cite{asynckv} & GPU attention comp $\leftrightarrow$ GPU KV prefetch (HBM$\rightarrow$L2) & KV block-wise \\
& KVPR~\cite{kvpr} & GPU KV re-comp $\leftrightarrow$ KV transfer (CPU$\leftrightarrow$GPU) & Split point \\
& CXL-SpecKV~\cite{cxl-speckv} & GPU comp $\leftrightarrow$ GPU KV prefetch (CXL memory$\rightarrow$L2) & KV page-wise \\
\midrule
I/O--Comm & PRESERVE~\cite{preserve} & GPU KV prefetch (HBM$\rightarrow$L2) $\leftrightarrow$ GPU collective comm & Operator-wise \\
\bottomrule
\end{tabular}
}
\end{table*}

\subsection{Pipelining and Overlapping}
\label{sec_po}

Pipelining and overlapping (\textit{OVLP}) methods hide KV-related latency by overlapping compute, I/O, and communication. Though often embedded in broader systems, 
%
Tab.~\ref{tab_overlap} highlights methods where OVLP forms the core technical contribution. \rev{We summarize them by mode and list the corresponding overlapped operations and granularity.} OVLP is key to reducing idle time and improving efficiency.



\subsection{Hardware-aware Execution}
\label{sec_hs}

This section focuses on hardware-aware execution (\textit{HAE}) methods that adapt KV-related operations to the underlying heterogeneous hardware.

\subsubsection{Disaggregated Inference}
\label{sec_di}



Disaggregated inference decouples inference compute onto distinct hardware resources to reduce contention and improve utilization. Infinite-LLM~\cite{infinitellm} adopts this idea at the operator level by splitting attention across distributed instances.
Several systems instead apply prefill-decode (PD) disaggregation, assigning compute-bound prefill and memory-bound decode to different compute pools~\cite{tetriinfer, splitwise, distserve, dejavu}.
%
Mooncake~\cite{mooncake} further couples PD disaggregation with a KV-centric scheduler and distributed cache pool, while
MuxServe~\cite{muxserve} colocates PD jobs within each GPU through SM partitioning.

\subsubsection{Compute Offloading}
\label{sec_comp_offload}

Compute offloading relocates partial compute to auxiliary devices to reduce GPU bottlenecks, utilizing hardware heterogeneity and workload features. 

A practical instantiation is \textbf{CPU offloading}, which leverages host CPUs for memory-intensive compute~\cite{fastdecode, cpucompv2, magicpig, neo}. They often follow a compute-near-cache principle for better locality. 
%
For example, FastDecode~\cite{fastdecode} and Neo~\cite{neo} offload both attention and KV caches, using cost-aware hardware selection and a load-aware scheduler, respectively. 

Beyond CPUs, several methods offload compute to \textbf{alternative devices}, such as computational storage drive (CSD)~\cite{instinfer} and processing-in-memory (PIM)~\cite{papi, attacc}. 
These methods expand the compute offloading space to broader device heterogeneity.


\begin{insightcallout}[]{\footnotesize\textbf{\rev{Takeaways \& Limitations -- Temporal Behavior}}}
\footnotesize
\rev{
\begin{itemize}[leftmargin=*,itemsep=2pt,topsep=1pt,parsep=0pt]
  \item KVS and OVLP directly target KV reuse and stall hiding. Lightweight cost models or predictors often enhance them. However, they are typically evaluated on controlled workloads, with limited analysis of robustness under bursty traffic or multi-tenant settings. 
  \item HAE improves throughput and hardware utilization by decoupling compute and specializing kernels, but its reliance on low-level primitives can make portability non-trivial for practitioners in some cases.
\end{itemize}
More analysis is provided in Apps.~\ref{a_takeaway_kvs_ovlp} and~\ref{a_takeaway_hae}. }
\end{insightcallout}

\section{KV Placement and Migration}
\label{sec_move}

\begin{figure}
    \centering
    \includegraphics[width=1.0\linewidth]{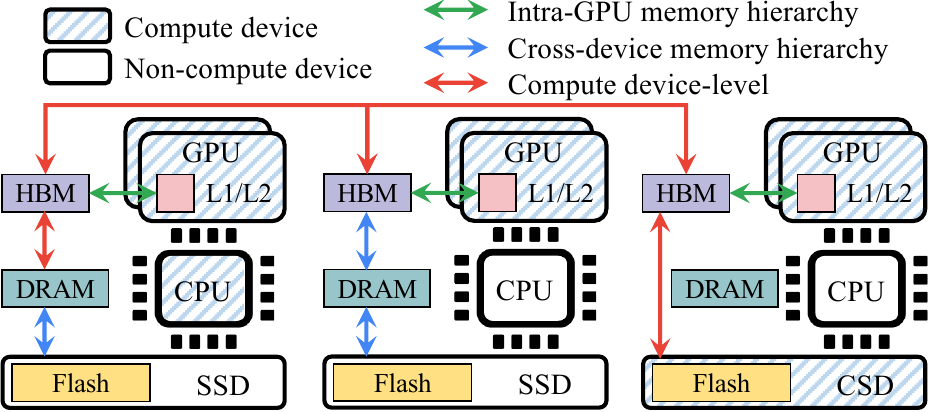}
    \vspace{-6mm}
    \caption{Illustration of KV cache placement and migration across memory hierarchies and compute devices.}
    \label{fig_mem}
    \vspace{-1mm}
\end{figure}

This section focuses on the spatial behaviors of how KV caches are placed and migrated across memory hierarchies and between compute devices. Figure~\ref{fig_mem} visualizes the architecture and transfer paths.

\subsection{Memory Hierarchy KV Orchestration}
\label{sec_hierarchy}


To scale under memory limits, we survey memory hierarchy KV orchestration (\textit{MHO}) methods that distribute KV caches across memory hierarchies.

\paragraph{Cross-device Memory Hierarchy.} 
A broad range of methods migrate KV entries across faster but limited GPU HBM, and larger but slower alternatives like CPU DRAM or SSD. 
Most of these works are \textbf{importance-aware}, designing importance scoring policies that maintain only critical KV entries on the GPU~\cite{alisa, infinigen, infllm, arkvale, impress, pensieve, omnikv, clusterkv, pqcache, shadowkv, specache, sliminfer, retrievalattention}. For instance, ArkVale~\cite{arkvale}, OmniKV~\cite{omnikv}, ClusterKV~\cite{clusterkv}, PQCache~\cite{pqcache}, and SpeCache~\cite{specache} offload the full KV cache to the CPU and keep only a lightweight proxy signal on the GPU. They then estimate importance via the proxy signal to guide the next prefetch.

Another line of cross-device methods optimizes KV placement and migration from a \textbf{system cost} view. FlexGen~\cite{flexgen} places KV caches across GPU, CPU, and disk via a cost model that maximizes throughput under bandwidth and latency constraints. Recent works further extend this hierarchy to compute express link (CXL)-based shared or disaggregated memory pools, enabling low-latency KV access and management~\citep{cxl-speckv, beluga, tract}. At runtime, many systems make online decisions about KV offloading or reloading based on system-level signals, such as queueing state, memory pressure, compute and I/O costs, and future reuse signals~\cite{fastserve, cachedattention, dejavu, layerkv, fastswitch, ragcache, lmcache, kvflow}.

\begin{table*}[t]
\centering
\footnotesize
\setlength{\tabcolsep}{2pt}
\caption{Summary of KV cache quantization (q.) methods. ``Avg. bits'' shows the average bitwidth per KV element \rev{based on the reported main results. This metric indicates memory savings and is comparable across methods.}}
\label{tab_kv_quant}
\vspace{-2mm}
\resizebox{\textwidth}{!}{
\begin{tabular}{lcccccc}
\toprule
\multirow{2}{*}{Method} & \multicolumn{2}{c}{Granularity} & \multirow{2}{*}{\makecell{Prec.\\mode}} & \multirow{2}{*}{\makecell{Important\\region}} & \multirow{2}{*}{\makecell{Outlier\\handling}} & \multirow{2}{*}{\makecell{Avg.\\bits}}\\
& Keys & Values &  &  &  \\
\midrule
SmoothQuant~\cite{smoothquant} & \multicolumn{2}{c}{Channel-wise} & Fixed & -- & Smoothing via scaling & 8 \\
FlexGen~\cite{flexgen} & \multicolumn{2}{c}{Group-wise} & Fixed & -- & -- & 4 \\
WKVQuant~\cite{wkvquant} & \multicolumn{2}{c}{2D (channel \& token)} & Mixed & Current token & Dynamic token-wise q. & \textasciitilde4 \\
MiKV~\cite{mikv} & \multicolumn{2}{c}{Token-wise} & Mixed & Existing policy & Outlier balancing & \textasciitilde4 \\
QAQ~\cite{qaq} & \multicolumn{2}{c}{Token-wise} & Mixed & Attention-aware & Sparse matrix (FP16) & 1.8-2.7 \\
Atom~\cite{atom} & \multicolumn{2}{c}{Group-wise} & Mixed & Outlier channels & Selective high-bits & 4.25 \\
KIVI~\cite{kivi} & Channel-wise & Token-wise & Mixed & Recent tokens & Channel-wise confining & \textasciitilde2 \\
CacheGen~\cite{cachegen} & \multicolumn{2}{c}{Layer-wise} & Mixed & Shallow layers & -- & 1.9-2.9 \\
DecoQuant~\cite{decoquant} & \multicolumn{2}{c}{Decomposed-tensor-wise} & Mixed & Small tensors & Tensor decomposition & 4 \\
GEAR~\cite{gear} & Channel-wise & Token-wise & Fixed & -- & Sparse matrix (FP16) & 4.4/5.0 \\
SKVQ~\cite{skvq} & \multicolumn{2}{c}{Group-wise} & Mixed & Init. \& recent tokens & Clipped dynamic q. & 2.25 \\
KVQuant~\cite{kvquant} & Channel-wise & Token-wise & Mixed & First token & Sparse matrix (FP16) & 4.3 \\
CQ~\cite{cq} & \multicolumn{2}{c}{Token-wise channel-group} & Fixed & -- & -- & 1.3 \\
ZipCache~\cite{zipcache} & Channel-wise & Chan.-sep. token-wise & Mixed & Norm. attention & Channel-wise norm. & 3.2 \\
QJL~\cite{qjl} & \multicolumn{2}{c}{Token-wise} &  Fixed & -- & Selective high-bits & 3/5 \\
VQ-LLM~\cite{vqllm} & \multicolumn{2}{c}{Group-wise (configurable)} & Fixed & -- & -- & 2/4 \\
SQuat~\cite{squat} & Block-wise & Token-wise & Fixed & -- & -- & 3.1 \\
QoQ~\cite{qserve} & \multicolumn{2}{c}{Channel-wise} & Fixed & -- & Smooth attention & 4 \\
CommVQ~\cite{commvq} & \multicolumn{2}{c}{Token-wise vector} & Fixed & -- & -- & 2 \\
OTT~\cite{ott} & Channel-wise & Token-wise & Mixed & Outlier \& recent tokens & Full precision & 2.5 \\
NSNQuant~\cite{nsnquant} & \multicolumn{2}{c}{Token-wise vector} & Fixed & -- & Token-wise norm. & 1.2/2.2 \\
ChanMix~\cite{chanmix} & Channel-wise & Token-wise & Mixed & Retrieval channels & Chan.-wise bit reallocation & 3.1 \\
\bottomrule
\end{tabular}
}
\vspace{-2mm}
\end{table*}

\paragraph{Intra-GPU Memory Hierarchy.}
Another line of MHO methods migrates KV entries between on-chip L1/L2 caches and off-chip HBM. \citet{asynckv} asynchronously prefetched upcoming KV blocks from HBM into L2 so that subsequent attention steps mostly hit in L2. Similarly, PRESERVE~\cite{preserve} fetches KV caches and inserts such operations selectively via graph-level optimization to avoid cache pollution.

\subsection{Compute Device KV Orchestration}
\label{sec_m}

Unlike hierarchical memory, compute device KV orchestration (\textit{CDO}) places and moves KV across compute-capable devices to enable distributed or heterogeneous serving. A common line performs intra-cluster orchestration, typically coupled with PD disaggregation~\cite{fastserve, splitwise, distserve, infinitellm, lmcache, tract}.
%
For instance, DistServe~\cite{distserve} proposes placement schemes for prefill and decode across high and low node-affinity GPU clusters, and uses a pull-based scheme where decode GPUs fetch KV as needed from prefill GPUs. TraCT~\citep{tract} exploits CXL shared memory as a substrate for KV transfer across disaggregated PD workers.

Beyond tightly coupled clusters, CacheGen~\cite{cachegen} targets remote KV transfer in network setups. It reduces network delay by encoding KV tensors into bitstreams and adaptively streaming them based on runtime bandwidth.
Finally, CDO also extends to heterogeneous accelerators, offloading attention to devices such as PIMs and CSDs~\cite{attacc, instinfer}.

\begin{insightcallout}[]{\footnotesize\textbf{\rev{Takeaways \& Limitations -- Spatial Behavior}}}
\footnotesize

\rev{MHO and CDO directly target interconnect bottlenecks through tiered KV management, and most systems overlap KV transfers with compute to hide latency, which is a central factor in their effectiveness. However, bandwidth} \rev{is typically handled without explicit modeling contention among concurrent KV transfers, making tail behavior hard to analyze. Another gap is the explicit joint optimization of offload and prefetch under shared memory and interconnect budgets. More takeaways are provided in App.~\ref{a_takeaway_spatial}.}
\end{insightcallout}



\section{KV Representation and Retention}
\label{sec_mem}


This section focuses on structural system behaviors of KV cache representation and retention.

\subsection{KV Cache Compression}
\label{sec_kvcc}

KV cache compression (\textit{KVCC}) is a central research thrust as it directly reduces memory usage.



\subsubsection{KV Cache Quantization}
\label{sec_kvq}

Quantization compresses floating-point tensors into lower-precision formats. Early works enable 8- and 4-bit KV~\cite{smoothquant, flexgen}. Later schemes adopt mixed precision, assigning high precision to critical KV entries. We compare methods \rev{along core algorithmic axes and effective bitwidths} in Tab.~\ref{tab_kv_quant} and present key insights here. 

\rev{One recurring pattern is \textbf{asymmetric KV quantization} (cf. ``granularity'' in Tab.~\ref{tab_kv_quant})}, as keys and values exhibit distinct outlier patterns and quantization sensitivities. For example, a common practice is to quantize keys per-channel and values per-token.
A second insight is that \textbf{outliers} play a crucial role in quantization, so many methods store them in higher bitwidths or design dedicated outlier handling techniques \rev{(cf. ``outlier handling'' in Tab.~\ref{tab_kv_quant})}.
These issues become more pronounced at very low bitwidths, where outliers and heterogeneous sensitivities are harder to preserve under such tight budgets. Thus, the compression-quality trade-off is an important practical consideration in this regime.





Recent advances use \textbf{vector quantization (VQ)} to compress groups with codebooks and capture inter-element correlation. CQ~\cite{cq} couples channels and learns centroids for 1-bit KV. VQ-LLM~\cite{vqllm} and CommVQ~\cite{commvq} reduce overhead via fused VQ kernels and RoPE-commutative codebooks. \citet{nsnquant} further improved calibration robustness.



It is worth noting that lower bitwidth does not always translate into end-to-end system gains. Realized system gains also depend on the runtime cost of (de)quantization, extra kernel boundaries, and whether these costs can be fused or overlapped. In practice, memory savings are more likely to yield latency or throughput gains when workloads are memory- or I/O-bound and quantization is  effectively integrated into the system stack.

\begin{table}[t]
\footnotesize
\centering
\setlength{\tabcolsep}{2pt}
\caption{\rev{Summary of low-rank approximation methods.}}
\label{tab_lr}
\vspace{-2mm}

\begin{threeparttable}
\rev{
\begin{adjustbox}{scale=0.91,center}
\begin{tabular}{ll@{\hspace{-10pt}}cl}
\toprule
Target & Method & Granularity & Rank \\
\midrule
\multirow{3}{*}{\makecell{Cached KV\\tensors}} & xKV~\cite{xkv} & LG & \tagF \\
& ReCalKV~\cite{recalkv} & K: HG; V: L & \tagB \\
& ShadowKV~\cite{shadowkv} & L (K-only) & \tagF \\
\midrule
\multirow{2}{*}{$W^{K}$,$W^{V}$} & LoRC~\cite{lorc} & L & \tagR \\
& Palu~\cite{palu} & HG & \tagS \\
\midrule
QKV & EigenAttention~\cite{eigenattention} & L & \tagB \\
\bottomrule
\end{tabular}
\end{adjustbox}
\begin{tablenotes}[flushleft]
\scriptsize
\item \textit{Granularity:} L = layer-wise; LG = layer-group-wise; HG = head-group-wise.
\item \textit{Rank policy:} \tagF\ fixed; \tagS\ searched; \tagB\ budget-driven; \tagR\ rule-based.
\end{tablenotes}
}
\end{threeparttable}
\vspace{-1.5mm}
\end{table}


\subsubsection{KV Cache Low-rank Approximation}
\label{sec_kvla}



%

Low-rank methods constrain KV-related tensors to a low-dimensional subspace, \rev{as summarized in Tab.~\ref{tab_lr} by target: (i) cached KV, (ii) KV projection weights $(W^K,W^V)$, or (iii) QKV attention subspace.} For instance, xKV~\cite{xkv} applies layer-group singular value decomposition to cached KV, while Palu~\cite{palu} factorizes $(W^K,W^V)$ with searched rank allocation.
KV tensor methods are most plug-and-play but add projection cost, whereas weight and QKV ones increase kernel coupling and engineering overhead. Some low-rank methods learn extra parameters, requiring training and thus out of scope under sKis.

\begin{table}[t]
\footnotesize
\centering
\setlength{\tabcolsep}{3pt}
\caption{\rev{Summary of structural compression methods.}}
\label{tab_struct}
\vspace{-2mm}

\begin{threeparttable}
\rev{
\begin{adjustbox}{scale=0.92,center}
\begin{tabular}{llcl}
\toprule
Family & Method & Unit & Signal \\
\midrule
\multirow{5}{*}{Merging} & KVMerger~\cite{kvmerger} & Token & \tagA\ \tagS \\
& CaM~\cite{cam}           & Token & \tagA \\
& KVSharer~\cite{kvsharer} & Layer & \tagS \\
& MiniCache~\cite{minicache} & Layer & \tagS \\
& D2O~\cite{d2o}           & Token & \tagS \\
\midrule
\multirow{3}{*}{Pruning} & CHAI~\cite{chai}  & Head & \tagS \\
& ThinK~\cite{think}       & Channel & \tagQ \\
 & ClusterAttn~\cite{clusterattn} & Token & \tagA \\
\bottomrule
\end{tabular}
\end{adjustbox}
\begin{tablenotes}[flushleft]
\scriptsize
\item \textit{Signal:} \tagA\ attention score; \tagS\ similarity/dissimilarity; \tagQ\ query norm.
\end{tablenotes}
}
\end{threeparttable}
\vspace{-1.5mm}
\end{table}

\begin{table*}[t]
\footnotesize
\centering
\setlength{\tabcolsep}{4pt}
\caption{\rev{Summary of representative KV cache eviction methods in chronological order.}}
\label{tab_evict}

\begin{threeparttable}
\resizebox{\textwidth}{!}{
\begin{tabular}{lclc}
\toprule
Method & Mode & Eviction policy & Budget policy \\
\midrule
H$_2$O~\cite{h2o} & Dynamic & \tagRR\ + H$_2$ (via accumulated attention) & Uniform  \\
Scissorhands~\cite{scissorhands} & Dynamic & \tagRR\ + Attention scores & Uniform  \\
RoCo~\cite{roco} & Dynamic & Mean \& std. dev. of attention scores & Uniform  \\
FastGen~\cite{fastgen} & Static & Hybrid (special/punctuation/locality/H$_2$) & Uniform \\
StreamingLLM~\cite{streamingllm} & Dynamic & \tagRR\ \tagS & Uniform  \\
Keyformer~\cite{keyformer} & Dynamic & \tagRR\ + Key (via Gumbel-softmax scores) & Uniform  \\
PyramidKV~\cite{pyramidkv} & Static & Observation window-based identification &  Preset (L, pyramid) \\
NACL~\cite{nacl} & Dynamic & Attention w.r.t. proxy tokens \& randomness & Uniform \\
PyramidInfer~\cite{pyramidinfer} & Dynamic & \tagRR\ + PvC (via ensemble attention) & Preset (L, pyramid) \\
BUZZ~\cite{buzz} & Dynamic & \tagRR\ \tagS\ + Segmented local H$_2$ & Uniform  \\
TOVA~\cite{tova} & Dynamic & Drop lowest attention score token at each step  &  Uniform \\
VATP~\cite{vatp} & Dynamic & \tagS\ + Attention \& value $L_1$-norm &  Uniform \\
L2KV~\cite{l2kv} & Dynamic & Key $L_2$-norm & Uniform \\
SnapKV~\cite{snapkv} & Static & Observation window-based identification & Uniform  \\
CAKE~\cite{cake} & Dynamic & \tagRR\ + Mean \& var. of attention scores & Adaptive (L, layer preference) \\
D$_2$O~\cite{d2o} & Dynamic & \tagRR\ \tagS\ + H$_2$ \& recall via merging (\S~\ref{sec_kvsc})& Adaptive (L, attention density) \\
SepLLM~\cite{sepllm} & Dynamic & \tagRR\ \tagS\ + Separator tokens & Uniform \\
LaCache~\cite{lacache} & Dynamic & Ladder pattern based & Preset (L, ladder) \\
KVCompose~\cite{kvcompose} & Dynamic & Aggregated attention \& form composite token & Adaptive (L,  composite importance) \\
DiffKV~\cite{diffkv} & Dynamic & \tagRR\ + Relative significance of attention scores & Adaptive (H, sparsity pattern)  \\
EvolKV~\cite{evolkv} & Dynamic & Plug-in (adopt existing eviction methods) &  Adaptive (L, evolutionary search) \\
DynamicKV~\cite{dynamickv} & Static & \tagRR\ + Attention w.r.t. instruction tokens & Adaptive (L, task-aware) \\
Ada-KV~\cite{adakv} & Dynamic & Plug-in (adopt existing eviction methods) &  Adaptive (H, attention sparsity) \\
\bottomrule
\end{tabular}
}
\begin{tablenotes}[flushleft]
\scriptsize
\item \textit{Eviction policy:} \tagRR\ recent tokens; \tagS\ attention sink tokens~\cite{streamingllm}, which means initial tokens. \hspace{0.4em} \textit{Budget policy:} L = layer-wise, H = head-wise.
\end{tablenotes}
\end{threeparttable}
\end{table*}

\subsubsection{KV Cache Structural Compression}
\label{sec_kvsc}

Unlike value-level methods, structural compression reduces KV memory by modifying cache organization \rev{(e.g., layer, head, channel, token). We compare existing methods in Tab.~\ref{tab_struct}, including (i) \textbf{pruning}, which drops a subset of structural units, and (ii) \textbf{merging}, which fuses units into shared forms. The decisions are often guided by attention or similarity measures (cf. ``signal'' in Tab.~\ref{tab_struct}), with clustering sometimes used to form groups~\cite{kvmerger, chai, clusterattn}.}

\subsection{KV Cache Retention Management}
\label{sec_clm}

Going beyond representations, this section focuses on mechanisms that efficiently manage the retention of the KV cache (\textit{KVRM}) during serving. 

\subsubsection{KV Cache Allocation and Reuse}
\label{sec_kvar}



\textbf{Structure-aware} methods redesign KV cache layouts for flexible allocation and reuse. One line targets virtualized allocation~\cite{vllm, vtensor, fastswitch, vattention}. A famous example is PagedAttention~\cite{vllm}, which uses fixed-size pages with logical-to-physical mapping to reduce fragmentation and support memory reuse. Another line builds structured indices for prompt sharing~\cite{promptcache, chunkattention, radixattention}, exemplified by radix tree~\cite{radixattention}. A third line standardizes KV layouts for kernels; \citet{flashinfer} introduced a block-sparse and composable format.



Orthogonally, \textbf{semantics-guided} methods further reduce materialization by computing KV only for critical tokens~\cite{lazyllm} and extend reuse to disaggregated LLM serving via an elastic MemPool system~\cite{memserve}.

\begin{table*}[tb]
\footnotesize
\centering
\setlength{\tabcolsep}{4.5pt}
\begin{threeparttable}
\caption{Behavior $\times$ objective matrix of sKis methods. Side bars encode research density (rows/columns). Cells mark relevance levels (\Strong = direct, \Weak = indirect) and high-prevalence flags (\istar: $\geq 70\%$ of papers report gains).}
\label{tab_behav-obj}
\vspace{-4mm}
\begin{tabularx}{\linewidth}{lcccccccc}
\toprule
\makecell[l]{Behaviors} &
\makecell{Mean\\latency} &
\makecell{Tail\\latency} &
\makecell{Through-\\put} &
\makecell{GPU\\memory} &
\makecell{Inter-\\connect I/O} &
\makecell{Energy\\/power} &
\makecell{Quality\\impact $\downarrow$} &
\makecell{Row\\density} \\
\midrule
\cellcolor{mypink!80}KV-centric scheduling & \Strong \istar & \Strong & \Strong & \Weak & \Weak & \Weak & \Weak & \heat{dense3} \\
\cellcolor{mypink!80}Pipelining and overlapping & \Strong \istar & \Weak & \Strong \istar &   & \Weak & \Weak & \Weak & \heat{dense4} \\
\cellcolor{mypink!80}Hardware-aware execution & \Strong \istar & \Strong & \Strong \istar &  & \Strong & \Strong & \Weak & \heat{dense3} \\
\cellcolor{mypurple!80}Memory hierarchy KV orchestration & \Weak & \Weak & \Weak & \Weak & \Strong \istar & \Strong & \Strong & \heat{dense2} \\
\cellcolor{mypurple!80}Compute device KV orchestration & \Weak & \Weak & \Strong \istar & \Weak & \Strong & \Strong & \Strong & \heat{dense4} \\
\cellcolor{mygreen!80}KV cache compression & \Weak & \Weak & \Weak & \Strong \istar & \Strong & \Strong & \Strong & \heat{dense1} \\
\cellcolor{mygreen!80}KV cache retention management & \Weak & \Weak & \Weak & \Strong \istar & \Strong & \Strong & \Strong & \heat{dense1}  \\
\midrule
Column density & \heat{dense1} & \heat{dense4} & \heat{dense2} & \heat{dense1} & \heat{dense3} & \heat{dense4} & \heat{dense1} & \\
\bottomrule
\end{tabularx}

\begin{tablenotes}
\item \TextLegend\hspace{0.2em}\RowDensityLegend\hspace{0.6em}\ColDensityLegend 
\end{tablenotes}
\end{threeparttable}
\end{table*}




\subsubsection{KV Cache Eviction}
\label{sec_kve}

KV cache eviction reduces memory by discarding less critical token KV states under a budget. \rev{We compare algorithmic details of existing methods in Tab.~\ref{tab_evict} and highlight three key insights.}

\rev{First, methods differ in when eviction is applied (cf. ``mode'' in Tab.~\ref{tab_evict}).} Static methods evict once during or after prefill and keep the retained set fixed in decoding, while dynamic ones update online during decoding to track importance shifts. \rev{Second, eviction policies often retain a recent window or attention sink tokens, and select extra tokens by lightweight signals such as attention-derived scores, heuristics, or robust variants that mitigate bias in local attention statistics (cf. ``eviction policy'' in Tab.~\ref{tab_evict}). Third, recent works move beyond uniform budgets and instead assign budgets across layers and even heads via preset and adaptive allocation (cf. ``budget policy'' in Tab.~\ref{tab_evict}).} Some methods treat budget policy as a plug-in to existing eviction rules.

\begin{insightcallout}[]{\footnotesize\textbf{Takeaways \& Limitations -- Structural Behavior}}
\footnotesize
\begin{itemize}[leftmargin=*,itemsep=2pt,topsep=1pt,parsep=0pt]
  \item KVCC delivers the most direct memory relief, but such savings may not translate into system gains without system co-design, due to tail (e.g., outlier) behavior, compression overhead, and kernel or runtime constraints. Practical deployment of KVCC remains a challenge.
  \item KVRM improves effective capacity by deciding which KV states exist at runtime. The key challenge is fast and stable utility estimation. In practice, policies are often workload-sensitive, and robustness under complex serving environments remains under-studied.
\end{itemize}
\rev{More analysis is provided in Apps.~\ref{a_takeaway_kvcc} and~\ref{a_takeaway_evict}.}
\end{insightcallout}


\section{Observations and Open Challenges}
\label{sec_observation_and_challenge}

In this section, we identify key observations from two complementary lenses: (i) a behavior $\times$ objective matrix and (ii) a behavior-behavior co-design affinity network, which naturally motivate open challenges of future sKis research. 


\subsection{Key Observations}
\label{sec_observation}

\begin{figure}
    \centering
    \includegraphics[width=1.0\linewidth]{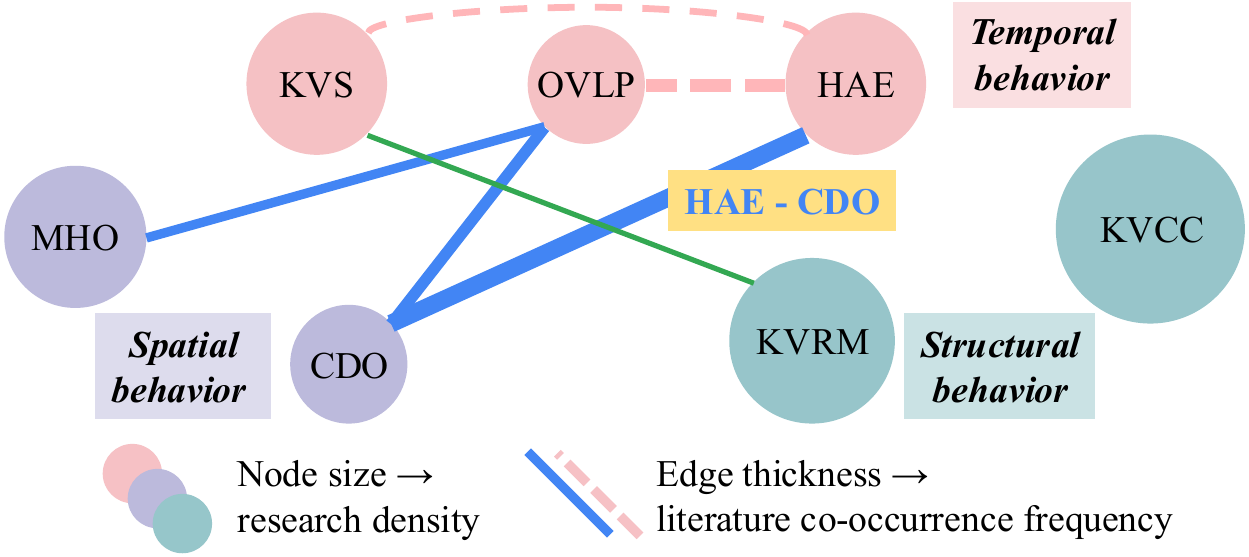}
    \vspace{-4mm}
    \caption{Behavior-behavior \rev{co-design affinity network}. }
    \label{fig_synergy}
\end{figure}

Table~\ref{tab_behav-obj} (behavior $\times$ objective matrix) marks each behavior's impact on serving objectives as direct (\Strong) or indirect (\Weak); stars (\istar) on direct cells statistically flag $\geq70\%$ of papers reporting such gains. Side bars show research density. Serving objectives cover latency, throughput, GPU memory, interconnect I/O, and energy. We also include quality impact $\downarrow$ to capture degradation as a trade-off. 
Figure~\ref{fig_synergy} (behavior-behavior \rev{co-design affinity} network) visualizes cross-behavior co-occurrence in the literature, with edge thickness proportional to normalized weights. Low-score edges are omitted, and computation details are provided in App.~\ref{a_synergy}). This affinity reflects observed co-design patterns rather than validated performance gains. 
We conclude key observations (\textbf{O1}-\textbf{O7}) as follows.

\noindent\textbf{O1. Structural works are most studied and dominate memory savings,} while others yield savings indirectly (e.g., via migration or reuse), indicating a community bias toward memory efficiency.

\noindent\textbf{O2. Temporal behaviors act most directly on latency and throughput,} since KVS, OVLP, and HAE map cleanly to reductions in scheduling stalls, pipeline bubbles, and device under-utilization. However, tail latency reporting is sparse.

\noindent\textbf{O3. Spatial methods primarily target interconnect I/O, often paired with OVLP.} Their core focus is KV transfer, and by overlapping it with compute, they effectively hide transfer latency.

\noindent\textbf{O4. Energy is under-explored,} although many methods reduce memory or compute intensity that should translate to energy benefits. 

\noindent\textbf{O5. Quality loss is universal.} Temporal methods risk inconsistent request handling; spatial methods risk missed KV data; and structural methods directly reduce KV precision. The practical question is to ensure such degradation is controllable.

\noindent\textbf{O6. HAE–CDO is the strongest \rev{co-design pattern}.}
Compute layouts that exploit device heterogeneity often co-design with KV colocating or transfer, yielding joint gains in utilization and I/O.


\noindent\textbf{O7. KVCC remains isolated} despite its popularity, which suggests a missed opportunity for co-design.

\subsection{Open Challenges}
\label{sec_challenge}

The above key observations (\textbf{O1}-\textbf{O7}) reveal both progress and gaps of current sKis research, which naturally motivate the next set of system-level open challenges (\textbf{C1}-\textbf{C6}).

\noindent\textbf{C1. SLO-driven tail control $\leftarrow$ O2.} Service-level objectives (SLOs) are critical to LLM serving, with tail latency dominating user experience~\cite{dean2013tail, wang2024revisiting}, yet most systems omit tail metrics. Under long contexts and bursts, KV generation, migration, and compression may trigger SLO violations, which often arise from cross-behavior interference rather than a single operation. The real challenge is to attribute these SLO violations to concrete KV behaviors and paths, motivating future sKis studies on standardized preemption and degradation semantics to make tail outcomes controllable.

\noindent\textbf{C2. Energy-aware sKis $\leftarrow$ O4.} With surging data center demand, AI infrastructure should be energy-aware~\cite{he2025a}, but energy is still rarely reported or optimized in sKis methods. Future research could integrate power profiling into runtime decisions, establish serving-time energy models, and jointly optimize energy-latency-quality under power constraints. Beyond aggregate power, energy should also be measured as per-request or per-token energy under serving traces and budgets. Another possible direction is to study energy-friendly KV granularities and layouts.

\noindent\textbf{C3. Trustworthy and efficient sKis $\leftarrow$ O5.} LLM serving must ensure not only quality but also trustworthiness~\cite{han2025trustworthy, yang2024regulating}, yet trust risks are rarely considered, leaving a gap between efficiency gains and trust failures. For example, structural methods can harm robustness in ways standard metrics miss, as policies may evict or compress low-salience but crucial context, causing severe errors under workload shifts despite stable mean accuracy. Notably, sKis techniques can be dual-use: \citet{robustkv} turned KV eviction into a defense against jailbreak attacks, suggesting that sKis techniques may become trust mechanisms. 
Trust concerns also span reliability, privacy, and safety across diverse sKis behaviors, as discussed in App.~\ref{a_trust}, and arise in various downstream settings~\cite{dongcarprt, chen2024benchmarking, jiang2026scribe}. Beyond average quality, future work should expose worst-case failures and quality lower bounds under stress, making trustworthiness behavior-attributed and SLO-aware.


\noindent\textbf{C4. Generalizable HAE–CDO $\leftarrow$ O6.} While HAE and CDO form the strongest co-design pattern, policies are often tailored to specific fabric or single-tenant settings. More broadly, although current sKis practice is largely GPU-centric, the temporal–spatial–structural lens itself is hardware-agnostic. What changes across hardware platforms is the concrete hierarchy, interconnect, and kernel or runtime support. Future directions include making such pattern portable across heterogeneous topologies (e.g., NVLink, NVSwitch, PCIe, CXL) and adaptive to multi-tenant settings.


\noindent\textbf{C5. Co-optimization and intermediate semantics $\leftarrow$ O7. } Most sKis optimize behaviors in isolation, despite their interactions under bandwidth and latency constraints. Future studies could explore co-optimization under shared budgets. For instance, to co-decide eviction, offload, and prefetch given predicted reuse, success probability, and I/O contention. In resource-constrained edge settings, this also suggests coupling KVCC with OVLP: KVCC reduces memory footprint, while OVLP helps prevent (de)compression overhead from negating system gains. Another promising direction is to exploit fine-grained intermediate semantics for behaviors and view co-optimization as state transitions over them. We give concrete examples in App.~\ref{a_intermediate} illustrating how intermediate semantics enable co-optimizing eviction, compression, and migration.



\noindent\textbf{C6. Unified benchmarks.} A fundamental challenge in sKis lies in establishing standardized benchmarks. We review LLM inference benchmarking practices in App.~\ref{a_benchmark_review}. We find inconsistent metric definitions and measures across tools, preventing reliable apples-to-apples comparisons across papers. We therefore advocate unified sKis benchmarks \rev{and offer a concise set of benchmark design principles, covering metrics (e.g., trust metrics and KV-centric resource metrics), representative stress workloads, and reporting standards, detailed in App.~\ref{a_benchmark_future}.}

\section{Conclusion}

This survey presents a systematic overview of sKis, offering a system behavior-oriented taxonomy covering temporal, spatial, and structural dimensions. By cross-analyzing behavior-objective impacts and behavior-behavior \rev{co-design patterns}, we reveal overlooked regions and open challenges. We hope this survey inspires continued exploration toward efficient and trustworthy LLM serving.


\section*{Limitations}

This paper offers a comprehensive review and summary of current methods in the area of system-aware KV cache optimization. However, given the extensive body of related work and the rapidly evolving nature of this research area, we may have overlooked some equally valuable contributions. We tried to include all relevant studies and references wherever feasible.

Additionally, this survey conducts no new experiments. Our claims synthesize results reported in public papers and open-source implementations, primarily under mainstream platforms and common configurations, which may constrain the generality of our conclusions. \rev{We avoid aggregating raw speedup or memory numbers across papers, because the reported gains are tightly coupled with model, hardware, workload, or baseline choices.}



\section*{Acknowledgments}

Feng Liu is supported by the ARC through grants DE240101089, LP240100101, DP230101540, and by the NSF\&CSIRO Responsible AI program under grant 2303037.
ChatGPT was used only for limited wording and grammar suggestions.

\bibliography{custom}

\appendix

\section{Preliminaries on LLMs}
\label{a_llm}


LLMs are built from stacked Transformer blocks, each with multi-head self-attention (MHSA) and feed-forward network (FFN). These blocks are sequential, where the output of one block serves as the input to the next.
For the $i$-th attention head, MHSA applies learned projections to the input features $X$ to get queries, keys, and values:
\begin{equation*}
  Q_i = XW^{Q_i}, K_i = XW^{K_i}, V_i = XW^{V_i}.
\end{equation*}
Then the self-attention operation is applied to each tuple ($Q_i$, $K_i$, $V_i$) and get the output of $Z_i$:
\begin{equation*}
  Z_i = \mathrm{attention}(Q_i, K_i, V_i) = \mathrm{softmax}(\frac{Q_i K_i^\top}{\sqrt{d_k}})V_i,
\end{equation*}
where $d_k$ is the dimension of the keys. Finally, outputs of all the attention heads are concatenated:
\begin{equation*}
    Z = \mathrm{concat}(Z_1, Z_2, ..., Z_h) W^O,
\end{equation*}
where $W^O$ is the trainable parameters. Following this, the output of MHSA is fed into the FFN module, which applies two linear transformations with a nonlinear activation (e.g., ReLU):
\begin{equation*}
    \mathrm{FFN}(x) = \mathrm{ReLU}(x W_1 + b_1) W_2 + b_2,
\end{equation*}
where $W_1$, $W_2$, $b_1$, and $b_2$ are learnable parameters of the FFN. These modules together enable contextualized autoregressive modeling in LLMs.

\section{Design of Our Taxonomy}
\label{a_taxonomy}

Our taxonomy follows a system behavior-oriented view of sKis introduced in \S~\ref{sec_pre} and respects the domain boundary. Specifically, we classify techniques by their operational impact along temporal, spatial, and structural dimensions. This behavior-oriented perspective follows established practice in machine learning systems research~\cite{xiao2018gandiva, rajbhandari2020zero, wang2021sensai, bnsl, qiu2024power, bnai, li2026sepprune} and aligns closely with how serving systems are actually built and optimized in practice, allowing diverse methods to be interpreted under a unified framework.

For example, many methods perform KV cache \emph{selection} by identifying tokens (i.e., KV entries) that are more or less important for future computation. In our taxonomy, we do not treat selection itself as a category. In contrast, we classify methods based on the system action taken after selection:
(i) If unimportant KV entries are permanently discarded to free GPU memory, the method is categorized as KV cache eviction (cf. \S~\ref{sec_kve}) under the structural dimension; (ii) If unimportant KV entries are offloaded to secondary storage (e.g., CPU RAM) for possible future retrieval and reload, the method falls under memory hierarchy KV orchestration (cf. \S~\ref{sec_hierarchy}) under the spatial dimension; and (iii) If the tokens are retained in GPU memory but excluded from computation, the method is considered token-level scheduling, which is categorized as KV-centric scheduling (cf. \S~\ref{sec_scheduling}) under the temporal dimension.

\begin{table*}
\centering
\footnotesize
\setlength{\tabcolsep}{3pt}
\renewcommand{\arraystretch}{1.15}
\caption{Comparison of scope and taxonomy with existing surveys related to efficient LLM inference or serving.}
\label{tab_a_related_work}
\vspace{-0.5mm}

\begin{tabularx}{\textwidth}{@{}lZZZZY@{}}
\toprule
Survey & KV-centric & Serving only & No retrain & System metrics & Organizing principle \\
\midrule
\citet{miao2023towards} &  & \cmark &  & \cmark & Algorithm-, system-level \\
\citet{yuan2024llm} &  & \cmark &  & \cmark & Optimization layer (parameter-, algorithm-, system-, hardware-level) \\
\citet{li2024llm} &  & \cmark & \cmark & \cmark & System component (KV cache and memory, computation, cloud deployment, emerging research fields) \\
\citet{zhou2024survey} &  & \cmark &  & \cmark & Optimization layer (data-, model-, system-level) \\
\citet{zhen2025taming} &  & \rev{\cmark} & \rev{\cmark} & \rev{\cmark} & \rev{Serving scale (instance-, cluster-level, emerging scenarios)} \\
\citet{bai2024beyond} &  &  &  & \cmark & Lifecycle (architecture design, pre-training, fine-tuning, inference, system design) \\
\citet{xu2024survey} &  &  &  & \cmark & Optimization layered (architecture, algorithm, systems) \\
\midrule
\citet{shi2024keep} & \cmark &  &  & \cmark & Lifecycle (training, deploy, post-training) \\
\citet{li2024survey} & \cmark & \cmark &  & \cmark & Optimization layer (token-, model-, system-level) \\
\citet{liu2025kv} & \cmark & \cmark &  & \cmark & KV compression types (selective token, quantization, attention compression, hybrid) \\
\midrule
This survey (sKis) & \cmark & \cmark & \cmark & \cmark & System behaviors (temporal, spatial, structural dimensions) \\
\bottomrule
\end{tabularx}
\end{table*}


\section{Related Surveys}
\label{a_related_work}

Several recent surveys have covered the areas of efficient LLM inference and serving. \rev{\citet{miao2023towards} explored both algorithmic innovations and system architectures for efficient LLM serving, \citet{yuan2024llm} analyzed LLM inference techniques through a Roofline-based framework, \citet{zhou2024survey} organized efficient LLM inference methods across data-, model-, and system-level optimizations, \citet{li2024llm} examined system-level enhancements for LLM inference serving, \citet{zhen2025taming} reviewed recent advances across different LLM serving scenarios, while \citet{bai2024beyond} and \cite{xu2024survey} focused on resource-efficient LLMs.} However, these \textbf{general surveys} typically treat KV cache optimization as a minor component within the broader pipelines.

In contrast, dedicated surveys that focus on the KV cache remain rare. \citet{shi2024keep} adopted a lifecycle-based taxonomy spanning training-stage, deploy-stage, and post-training optimizations. \citet{li2024survey} categorized KV cache management strategies into token-level, model-level, and system-level optimizations. \citet{liu2025kv} focused on compression strategies of the KV cache, such as selective token strategies, quantization, and attention compression.
These \textbf{KV-specific surveys} are closest to our topic. However, they mostly organize by lifecycle stages or abstraction levels, leaving the serving-time behaviors of KV caches unexamined.

Different from the above surveys, we concentrate exclusively on the sKis scope (i.e., serving-time, KV-centric, system metrics, no retraining or architecture change) and aim to provide a deeper understanding within this scope. By classifying methods according to their impact along temporal, spatial, and structural dimensions, our survey enables cross-behavior and behavior$\times$objective analysis, which complements prior surveys and clarifies actionable research gaps. Table~\ref{tab_a_related_work} shows a summary.

\section{Supplementary Paper Categorization}
\label{a_extended_c}

\begin{table*}
\small
\centering
\begin{threeparttable}
\caption{Full mapping of representative methods reviewed in this paper to their corresponding sKis categories. Methods are chronologically ordered with publication venues.}
\label{tab_a_extended_c}
\vspace{-0.6mm}

\begin{tabularx}{\textwidth}{lYYYYYYYY}
\toprule
Methods & Venue & \multicolumn{7}{c}{Taxonomy of sKis} \\
\midrule
 & &
\rot{KV-centric scheduling} &
\rot{Pipelining and overlapping} &
\rot{Hardware-aware execution} &
\rot{Memory hierarchy KV orchestration} &
\rot{Compute device KV orchestration} &
\rot{KV cache compression} &
\rot{KV cache retention management} \\
\midrule
SmoothQuant~\cite{smoothquant} 
& ICML &  &  &  &  &  & \primary &  \\
FlexGen~\cite{flexgen} 
& ICML &  & \secondary & \secondary & \primary &  & \primary &  \\
vLLM~\cite{vllm} 
& SOSP &  &  &  & \secondary &  &  & \primary \\
FastServe~\cite{fastserve} 
& NeurIPS &  & \secondary &  & \primary & \primary &  &  \\
H$_2$O~\cite{h2o} 
& NeurIPS &  &  &  &  &  &  & \primary \\
Scissorhands~\cite{scissorhands} 
& NeurIPS &  &  &  &  &  &  & \primary \\
TetriInfer~\cite{tetriinfer} 
& arXiv & \primary &  & \primary &  & \secondary &  &  \\
RoCo~\cite{roco} 
& arXiv &  &  &  &  &  &  & \primary \\
WKVQuant~\cite{wkvquant} 
& arXiv &  &  &  &  &  & \primary &  \\
MiKV~\cite{mikv} 
& arXiv &  &  &  &  &  & \primary &  \\
FastDecode~\cite{fastdecode} 
& arXiv &  & \primary & \primary &  & \secondary & \secondary &  \\
QAQ~\cite{qaq} 
& arXiv &  &  &  &  &  & \primary &  \\
AttAcc~\cite{attacc} 
& ASPLOS &  &  & \primary &  & \primary &  &  \\
FastGen~\cite{fastgen} 
& ICLR &  &  &  &  &  &  & \primary \\
StreamingLLM~\cite{streamingllm} 
& ICLR &  &  &  &  &  &  & \primary \\
Preble~\cite{preble} 
& ICLR & \primary &  & \secondary &  &  &  & \secondary \\
Keyformer~\cite{keyformer} 
& MLSys &  &  &  &  &  &  & \primary \\
Atom~\cite{atom} 
& MLSys &  &  &  &  &  & \primary &  \\
PromptCache~\cite{promptcache} 
& MLSys &  &  &  &  &  &  & \primary \\
PyramidKV~\cite{pyramidkv} 
& arXiv &  &  &  &  &  &  & \primary \\
Splitwise~\cite{splitwise} 
& ISCA &  & \secondary & \primary &  & \primary &  &  \\
ALISA~\cite{alisa} 
& ISCA &  &  &  & \primary &  & \secondary & \secondary \\
DistServe~\cite{distserve} 
& OSDI &  & \secondary & \primary &  & \primary &  &  \\
Infinite-LLM~\cite{infinitellm} 
& arXiv &  & \secondary & \primary &  & \primary &  &  \\
InfiniGen~\cite{infinigen} 
& OSDI &  &  &  & \primary &  &  &  \\
CachedAttention~\cite{cachedattention} 
& ATC & \secondary & \primary &  & \primary & \secondary &  & \secondary \\
LazyLLM~\cite{lazyllm} 
& arXiv &  &  &  &  &  &  & \primary \\
KVMerger~\cite{kvmerger} 
& arXiv &  &  &  &  &  & \primary &  \\
vTensor~\cite{vtensor} 
& arXiv &  &  &  &  &  &  & \primary \\
KIVI~\cite{kivi} 
& ICML &  &  &  &  &  & \primary &  \\
CHAI~\cite{chai} 
& ICML &  &  &  &  &  & \primary &  \\
CaM~\cite{cam} 
& ICML &  &  &  &  &  & \primary &  \\
MuxServe~\cite{muxserve} 
& ICML & \primary &  & \primary &  & \secondary &  & \secondary \\
Quest~\cite{quest} 
& ICML & \primary &  &  &  &  &  &  \\
SparQAttention~\cite{sparq} 
& ICML & \primary &  &  &  &  &  &  \\
D{\'e}j{\`a}Vu~\cite{dejavu} 
& ICML &  &  & \primary & \primary & \secondary &  & \secondary \\
CacheGen~\cite{cachegen} 
& SIGCOMM &  & \secondary &  &  & \primary & \primary &  \\
DecoQuant~\cite{decoquant} 
& ACL &  &  &  &  &  & \primary &  \\
NACL~\cite{nacl} 
& ACL &  &  &  &  &  &  & \primary \\
PyramidInfer~\cite{pyramidinfer} 
& ACL &  &  &  &  &  &  & \primary \\
ChunkAttention~\cite{chunkattention} 
& ACL &  &  &  &  &  &  & \primary \\
InstInfer~\cite{instinfer} 
& arXiv &  & \secondary & \primary &  & \primary &  & \secondary \\
TwinPilots~\cite{twinpilots} 
& SYSTOR &  & \secondary & \primary &  & \secondary &  &  \\
GEAR~\cite{gear} 
& arXiv &  &  &  &  &  & \primary &  \\
LoRC~\cite{lorc} 
& arXiv &  &  &  &  &  & \primary &  \\
SKVQ~\cite{skvq} 
& COLM &  &  &  &  &  & \primary &  \\
LayerKV~\cite{layerkv} 
& arXiv & \primary & \secondary &  & \primary &  &  & \secondary \\
CComp~\cite{cpucompv2} 
& PACT &  & \primary & \primary &  & \secondary &  &  \\
KVSharer~\cite{kvsharer} 
& arXiv &  &  &  &  &  & \primary &  \\
LAMPS~\cite{lamps} 
& arXiv & \primary &  &  & \secondary &  &  & \secondary \\
BUZZ~\cite{buzz} 
& arXiv &  &  &  &  &  &  & \primary \\
LoongServe~\cite{loongserve} 
& SOSP & \primary & \secondary & \secondary &  &  &  & \secondary \\
EigenAttention~\cite{eigenattention} 
& EMNLP &  &  &  &  &  & \primary &  \\
TOVA~\cite{tova} 
& EMNLP &  &  &  &  &  &  & \primary \\
VATP~\cite{vatp} 
& EMNLP &  &  &  &  &  &  & \primary \\
L2KV~\cite{l2kv} 
& EMNLP &  &  &  &  &  &  & \primary \\
FastSwitch~\cite{fastswitch} 
& arXiv & \secondary &  & \secondary & \primary &  &  & \primary \\
\bottomrule
\end{tabularx}

\vspace{0.5em}
\begin{flushright}
\textit{Continued on next page}
\end{flushright}

\begin{tablenotes}
\item \primary~= primary category with main analysis in the paper; \secondary~= secondary category omitted or only briefly mentioned in our paper to maintain focused classification.
\end{tablenotes}
\end{threeparttable}
\end{table*}

\begin{table*}
\small
\centering
\begin{threeparttable}
\captionsetup{labelformat=empty, justification=raggedright, singlelinecheck=false}
\caption{\small\itshape Continued from previous page}
\vspace{-3.2mm}

\begin{tabularx}{\textwidth}{lYYYYYYYY}
\toprule
Methods & Venue & \multicolumn{7}{c}{Taxonomy of sKis} \\
\midrule
 & &
\rot{KV-centric scheduling} &
\rot{Pipelining and overlapping} &
\rot{Hardware-aware execution} &
\rot{Memory hierarchy KV orchestration} &
\rot{Compute device KV orchestration} &
\rot{KV cache compression} &
\rot{KV cache retention management} \\
\midrule
KVQuant~\cite{kvquant} 
& NeurIPS &  &  &  &  &  & \primary &  \\
CQ~\cite{cq} 
& NeurIPS &  &  &  &  &  & \primary &  \\
ZipCache~\cite{zipcache} 
& NeurIPS &  &  &  &  &  & \primary &  \\
SnapKV~\cite{snapkv} 
& NeurIPS &  &  &  &  &  &  & \primary \\
MiniCache~\cite{minicache} 
& NeurIPS &  &  &  &  &  & \primary &  \\
InfLLM~\cite{infllm} 
& NeurIPS &  &  &  & \primary &  &  & \secondary \\
RadixAttention~\cite{radixattention} 
& NeurIPS & \primary &  &  &  &  &  & \primary \\
Loki~\cite{loki} 
& NeurIPS & \primary &  &  &  &  &  &  \\
ArkVale~\cite{arkvale} 
& NeurIPS &  &  &  & \primary &  & \secondary &  \\
MemServe~\cite{memserve} 
& arXiv &  &  &  &  &  &  & \primary \\
Mooncake~\cite{mooncake} 
& FAST & \primary & \secondary & \primary &  & \secondary &  & \\
IMPRESS~\cite{impress} 
& FAST &  &  &  & \primary &  &  & \secondary \\
QJL~\cite{qjl} 
& AAAI &  &  &  &  &  & \primary &  \\
VQ-LLM~\cite{vqllm} 
& HPCA &  &  &  &  &  & \primary &  \\
xKV~\cite{xkv} 
& arXiv &  &  &  &  &  & \primary &  \\
SQuat~\cite{squat} 
& arXiv &  &  &  &  &  & \primary &  \\
vAttention~\cite{vattention} 
& ASPLOS &  & \secondary &  &  &  &  & \primary \\
PAPI~\cite{papi} 
& ASPLOS &  &  & \primary &  &  &  &  \\
Pensieve~\cite{pensieve} 
& EuroSys & \secondary & \secondary &  & \primary &  &  & \secondary \\
AsyncKV~\cite{asynckv} 
& arXiv &  & \primary & \secondary & \primary &  &  &  \\
Palu~\cite{palu} 
& ICLR &  &  &  &  &  & \primary &  \\
CAKE~\cite{cake} 
& ICLR &  &  &  &  &  &  & \primary \\
D$_2$O~\cite{d2o} 
& ICLR &  &  &  &  &  & \primary & \primary \\
ThinK~\cite{think} 
& ICLR &  &  &  &  &  & \primary &  \\
MagicPIG~\cite{magicpig} 
& ICLR &  &  & \primary &  &  &  &  \\
OmniKV~\cite{omnikv} 
& ICLR & \secondary &  &  & \primary &  &  &  \\
QoQ~\cite{qserve} 
& MLSys &  &  &  &  &  & \primary &  \\
FlashInfer~\cite{flashinfer} 
& MLSys & \primary &  & \secondary & \secondary &  &  & \primary \\
Neo~\cite{neo} 
& MLSys &  & \primary & \primary &  & \secondary &  &  \\
PRESERVE~\cite{preserve} 
& arXiv &  & \primary & \secondary & \primary &  &  &  \\
ReCalKV~\cite{recalkv} 
& arXiv &  &  &  &  &  & \primary &  \\
ClusterKV~\cite{clusterkv} 
& DAC &  &  &  & \primary &  &  &  \\
PQCache~\cite{pqcache} 
& SIGMOD &  & \secondary &  & \primary &  & \secondary &  \\
ShadowKV~\cite{shadowkv} 
& ICML &  & \secondary &  & \primary &  & \primary &  \\
SepLLM~\cite{sepllm} 
& ICML &  &  &  &  &  &  & \primary \\
CommVQ~\cite{commvq} 
& ICML &  &  &  &  &  & \primary &  \\
LaCache~\cite{lacache} 
& ICML &  &  &  &  &  &  & \primary \\
SpeCache~\cite{specache} 
& ICML &  & \secondary &  & \primary &  &  &  \\
RocketKV~\cite{rocketkv} 
& ICML & \primary &  &  &  &  &  & \secondary \\
ClusterAttn~\cite{clusterattn} 
& ACL &  &  &  &  &  & \primary &  \\
RefreshKV~\cite{refreshkv} 
& ACL & \primary &  &  &  &  &  &  \\
OTT~\cite{ott} 
& ACL &  &  &  &  &  & \primary &  \\
KVPR~\cite{kvpr} 
& ACL &  & \primary & \secondary &  &  & \secondary &  \\
SlimInfer~\cite{sliminfer} 
& arXiv &  & \secondary &  & \primary &  &  &  \\
RAGCache~\cite{ragcache} 
& TOCS &  & \secondary &  & \primary &  &  &  \\
KVCompose~\cite{kvcompose} 
& arXiv &  &  &  &  &  &  & \primary \\
LMCache~\cite{lmcache} 
& arXiv &  & \secondary & \secondary & \primary & \primary &  & \secondary \\
DiffKV~\cite{diffkv} 
& SOSP &  &  &  &  &  & \secondary & \primary \\
TokenSelect~\cite{tokenselect} 
& EMNLP & \primary &  &  &  &  &  & \secondary \\
EvolKV~\cite{evolkv} 
& EMNLP &  &  &  &  &  &  & \primary \\
DynamicKV~\cite{dynamickv} 
& EMNLP &  &  &  &  &  &  & \primary \\
RetrievalAttention~\cite{retrievalattention} 
& NeurIPS &  &  &  & \primary &  &  &  \\
Ada-KV~\cite{adakv} 
& NeurIPS &  &  &  &  &  &  & \primary \\
NSNQuant~\cite{nsnquant} 
& NeurIPS &  &  &  &  &  & \primary &  \\
KVFlow~\cite{kvflow} 
& NeurIPS & \secondary & \secondary &  & \primary &  & \primary &  \\
TraCT~\cite{tract} 
& arXiv &  &  & \secondary & \primary & \primary &  &  \\
CXL-SpecKV~\cite{cxl-speckv} 
& FPGA &  & \primary & \secondary & \primary &  & \secondary &  \\
ChanMix~\cite{chanmix} 
& ICLR &  &  &  &  &  & \primary &  \\
Beluga~\cite{beluga} 
& SIGMOD &  &  &  & \primary &  &  &  \\
\bottomrule
\end{tabularx}

\vspace{0.5em}
\begin{flushright}
\textit{Continued on next page}
\end{flushright}

\begin{tablenotes}
\item \primary~= primary category with main analysis in the paper; \secondary~= secondary category omitted or only briefly mentioned in our paper to maintain focused classification.
\end{tablenotes}
\end{threeparttable}
\end{table*}





Table~\ref{tab_a_extended_c} provides a supplementary mapping of all surveyed methods across the full taxonomy of 7 subcategories under 3 major optimization dimensions. The finer-grained categories in this table include (i) KV-centric scheduling (cf. \S~\ref{sec_scheduling}), (ii) pipelining and overlapping (cf. \S~\ref{sec_po}), (iii) hardware-aware execution (cf. \S~\ref{sec_hs}), (iv) memory hierarchy KV orchestration (cf. \S~\ref{sec_hierarchy}), (v) compute device KV orchestration (cf. \S~\ref{sec_m}), (vi) KV cache compression (cf. \S~\ref{sec_kvcc}), and (vii) KV cache retention management (cf. \S~\ref{sec_clm}).

As discussed in \S~\ref{sec_pre}, each method is primarily discussed under one or two key optimization categories that reflect its main contributions. These categories are denoted as primary category (\primary) in Tab.~\ref{tab_a_extended_c}.
However, some methods also touch upon additional optimization aspects that are not covered or elaborated in the main sections. 
For example, to support its ``hardware-aware execution'' design of decoupling prefill and decode phases across heterogeneous devices, Splitwise~\cite{splitwise} incorporates a fine-grained layer-wise transmission strategy that transmits the KV cache from the prefill node to the decode node and overlaps such KV cache transmission with the computation in the prefill phase. They serve as enabling mechanisms that make the decoupled strategy feasible and link Splitwise to the ``device-level KV transfer'' and ``pipelining and overlapping'' categories.
We summarize these omitted associations in Tab.~\ref{tab_a_extended_c}, denoted by \secondary, to provide a more complete mapping for readers interested in cross-cutting techniques.


\paragraph{Venue Diversity.}
The sKis methods span a broad range of research communities as shown in Tab.~\ref{tab_a_extended_c}. The publication venues include top-tier machine learning and artificial intelligence conferences (e.g., ICLR, ICML, NeurIPS, AAAI), natural language processing venues (e.g., ACL, EMNLP, COLM), systems and architecture conferences (e.g., ASPLOS, ISCA, HPCA, FAST, ATC, EuroSys, OSDI, SOSP, SC, SIGCOMM, DAC, FPGA), and interdisciplinary forums such as MLSys and SIGMOD. We also include some impactful arXiv preprints. This diversity underscores the inherently cross-cutting nature of KV cache optimization and highlights the growing recognition of this topic across various research communities.

\section{Takeaways}
\label{a_takeaway}

Through a comprehensive literature review of sKis, we discovered takeaways across several domains. 

\subsection{Scheduling and Overlapping}
\label{a_takeaway_kvs_ovlp}

KVS and OVLP directly target runtime stalls. KVS prioritizes limited resources for the most reusable and latency-sensitive work; OVLP is also a type of scheduling, aligning compute with data transfer to fill pipeline bubbles. 

\noindent\faKey \textbf{Takeaway:}
\begin{itemize}[label=\cmark, leftmargin=*,itemsep=2pt,topsep=1pt,parsep=0pt]
  \item KVS is a multi-objective optimization problem. Modern schedulers often prioritize KV usage over time rather than FLOPS, and KV reuse-driven scheduling is the default paradigm. 
  \item KVS is enhanced by prediction. Lightweight predictors plus a robust policy outperform traditional FCFS or SJF schemes~\cite{tetriinfer, mooncake, lamps}.
  \item The key to OVLP is to perform at the true bottleneck with asymmetric pipelines. For example, keep compute-bound prefill on GPU, and overlap memory-bound decode attention and KV with I/O or collective communication. 
  \item Preferring recompute to transfer~\cite{kvpr}, or prefetching KV caches into L2 during collectives~\cite{asynckv, preserve}, can substantially reduce pipeline bubbles, especially when bandwidth is the bottleneck.
\end{itemize}

\subsection{Hardware-aware Execution}
\label{a_takeaway_hae}

HAE improves throughput, reduces mean/tail latency, and extends servable context without retraining, by decoupling phases and mapping execution to hardware capabilities.

\noindent\faKey \textbf{Takeaway:}
\begin{itemize}[label=\cmark, leftmargin=*,itemsep=2pt,topsep=1pt,parsep=0pt]
  \item Compute should follow hardware capabilities. When executing on a given device, it is critical to specialize kernels, tiling, and memory layouts to that device.
  \item Create KV locality within the device rather than moving KV across devices. It is effective to keep hot KV caches close to the compute.
  \item Compute-intensive prefill and memory-bound decode benefit from phase-specific execution mappings (cf. \S~\ref{sec_comp_offload}).
  \item HAE should adapt to the access granularity and parallelism of the target device.
\end{itemize}

\subsection{Placement and Migration}
\label{a_takeaway_spatial}

MHO and CDO govern where KV caches reside across the memory hierarchy and how they transfer during serving. They act directly on interconnect bandwidth bottlenecks, with GPU memory relief emerging as a by-product of tiering and offloading.

\noindent\faKey \textbf{Takeaway:}
\begin{itemize}[label=\cmark, leftmargin=*,itemsep=2pt,topsep=1pt,parsep=0pt]
  \item It is a common MHO pattern to keep only future-useful KV caches on the GPU, demote the rest to CPU or SSD, and reload guided by attention cues. Cost models can be effectively used to choose CPU, GPU, SSD paths~\cite{flexgen, ragcache}.
  \item Most MHO and CDO solutions overlap I/O transfers with compute or collectives to hide latency, although they often serve OVLP as a secondary category.
  \item Under interconnect bottlenecks, co-adaptation of transfer paths, precisions, or decoding strategies can reduce TTFT and SLO violations compared with static schemes~\cite{distserve, cachegen, fastswitch}.
  \item Migration granularity and path should align with attention access patterns and device access units.
  \item Prefetch-evict co-design is rare. The field would benefit from a unified objective that jointly accounts for prefetch deadlines and eviction risk.
\end{itemize}

\subsection{KV Cache Compression}
\label{a_takeaway_kvcc}

KV caches can quickly overwhelm the memory capacity of GPUs and pose bandwidth pressure as context length or batch size increases, since the size of the KV cache scales linearly with these two factors. Consequently, prior works have proposed various approaches to directly compress the KV cache, such as quantization, low-rank approximation, and structural compression.

\noindent\faKey \textbf{Takeaway:}
\begin{itemize}[label=\cmark, leftmargin=*,itemsep=2pt,topsep=1pt,parsep=0pt]
  \item Outliers dominate performance at low bitwidths or ranks~\cite{ott}. Isolating outliers (e.g., higher bitwidths) for value-level compression methods prevents worst-case error explosions.
  \item Recent advances trend toward applying vector quantization (VQ)~\cite{vq} for KV cache quantization, and they often reach very low-bit (i.e., 1-2 bits) quantization with modest quality loss~\cite{cq, vqllm, nsnquant, commvq}. 
  \item KVCC has been developed mostly at the algorithm level, while system-level integration is thin (cf. Fig.~\ref{fig_synergy}). Thus, memory reductions often fail to translate into lower mean/tail latency or higher throughput unless KVCC is co-designed with execution, migration, and runtime control, which is consistent with recent observations in DeltaKV~\cite{hao2026deltakv}.
\end{itemize}

We further discuss the co-design of KVCC with execution, migration, and runtime control (the last akeaway) as follows:
(i) Co-design with execution: quantization/de-quantization and low-rank updates can be fused into attention kernels or overlapped with compute, so compression overhead does not re-introduce stalls in the decode pipeline;
(ii) Co-design with migration: aligning compressed packing units with device access units ensures that memory footprint reductions translate into fewer, fully utilized transfer chunks that fit overlap windows. 
(iii) Co-design with runtime control: exposing tunable parameters (e.g., bitwidth, rank, sparsity) to the runtime and adjusting them under SLOs remains an opportunity beyond static configurations.



\subsection{KV Cache Eviction}
\label{a_takeaway_evict}

KV cache eviction decides which past tokens remain resident under tight memory and bandwidth budgets, so that long contexts can be served. It operates in both phases and trades memory and transfer cost against utility to the attention compute. 

\noindent\faKey \textbf{Takeaway:}
\begin{itemize}[label=\cmark, leftmargin=*,itemsep=2pt,topsep=1pt,parsep=0pt]
  \item KV cache eviction is important in both prefill and decode. The former focuses on the KV cache to be computed, while the latter focuses on the KV cache that has been computed.
  \item Most systems retain a small recent window, a tiny set of ``attention sink'' anchor tokens~\cite{streamingllm, guattention}, and a few ``heavy hitters''~\cite{h2o}.
  \item Token importance should not be judged by attention scores alone. KV norms provide strong and low-overhead signals~\cite{vatp, l2kv}. We also note broader links to interpretability work on importance scoring~\cite{ig, yang2023local, yang2023re}.
  \item It is effective to use heterogeneous budgets across layers or heads, rather than a uniform upper bound, as shown in Tab.~\ref{tab_evict}. For example, shallow layers often deserve larger retention, while deeper layers emphasize global semantics and tolerate more sparsity. 
  \item Pairing KV cache eviction with similarity-based recall or merge is stronger than hard deletion, preserving salient context under tight budgets and improving long-context consistency~\cite{d2o}.
\end{itemize}





\section{Behavior-behavior \rev{Co-design Affinity} Computation}
\label{a_synergy}


Below we detail the compute procedure of behavior pairs' normalized co-occurrence strengths, which are reflected by the edge thicknesses in Fig.~\ref{fig_synergy}.

Let $\mathcal{B}=$\{KVS, OVLP, HAE, MHO, CDO, KVCC, KVRM\} denote the set of behaviors and $\mathcal{P}$ the set of papers. For $p\in\mathcal{P}$ and $i\in\mathcal{B}$, let the categorical label be
$\ell_{p,i}\in\{\mathsf{P},\mathsf{S},\mathsf{NA}\}$, which means primary category (\primary), secondary category (\secondary), or no category. Each $\ell_{p,i}$ can be observed from Tab.~\ref{tab_a_extended_c}. We map labels to numeric weights by $\omega(\ell_{p,i}) = \mathbbm{1}_{[\ell_{p,i}=\mathsf{P}]} + \alpha\mathbbm{1}_{[\ell_{p,i}=\mathsf{S}]}$ with $\alpha=0.5$, where $\mathbbm{1}_{[\cdot]}$ is the indicator function that equals 1 when the stated condition holds and 0 otherwise.

\begin{table}[t]
  \centering
  \footnotesize
  \setlength{\tabcolsep}{2.5pt}
  \begin{tabular}{lccccccc}
    \toprule
     & KVS & OVLP & HAE & MHO & CDO & KVCC & KVRM \\
    \midrule
    KVS  & -- & 2.5 & 4.75 & 4 & 1.75 & 0 & 6.5 \\
    OVLP & 2.5 & -- & 9.25 & 8.5 & 6.5 & 2.75 & 2.25 \\
    HAE  & 4.75 & 9.25 & -- & 3.75 & 10 & 1.25 & 3.25 \\
    MHO  & 4 & 8.5 & 3.75 & -- & 3 & 3.5 & 5.75 \\
    CDO  & 1.75 & 6.5 & 10 & 3 & -- & 1.25 & 2.75 \\
    KVCC & 0 & 2.75 & 1.25 & 3.5 & 1.25 & -- & 1.75 \\
    KVRM & 6.5 & 2.25 & 3.25 & 5.75 & 2.75 & 1.75 & -- \\
    \bottomrule
  \end{tabular}
  \caption{Raw (pre-normalization) co-occurrence matrix that encodes the weighted co-occurrence strength between system behaviors across papers.}
  \label{tab_raw_C}
\end{table}

\paragraph{Constructing raw co-occurrence.}
The raw co-occurrence matrix $C\in\mathbb{R}^{|\mathcal{B}|\times |\mathcal{B}|}$ aggregates pairwise co-appearance strength, as shown in Tab.~\ref{tab_raw_C}. Each cell $C_{ij}$ of the behavior pair $i,j$ is defined by summing the per-paper products of their weights:
\begin{equation*}
    C_{ij} = \sum_{p\in\mathcal{P}} \omega(\ell_{p,i}) \omega(\ell_{p,j}).
\end{equation*}
%

\paragraph{Constructing normalized \rev{co-design affinity}.}
While the raw co-occurrence matrix $C$ captures absolute overlap, it is biased by marginal popularity, because the behaviors with larger research density tend to have larger $C_{ij}$. We therefore normalize $C$ using the Tanimoto coefficient.
We define the per-behavior squared weight $Q_i$:
\begin{equation*}
    Q_i = \sum_{p\in\mathcal{P}} w_{p,i}^2.
\end{equation*}
Then the Tanimoto-normalized \rev{co-design affinity} matrix $S\in\mathbb{R}^{|\mathcal{B}|\times |\mathcal{B}|}$ reflects relative co-occurrence strength on a $[0,1]$ scale, as shown in Tab.~\ref{tab_norm_S}. Each cell in $S_{ij}$ of the behavior pair $i,j$ is defined as the ratio of their shared weighted presence to their squared union:
\begin{equation*}
    S_{ij} = \frac{C_{ij}}{\,Q_i+Q_j-C_{ij}\,}.
\end{equation*}
Compared to $C_{ij}$, this score controls marginal sizes and is visualized in Fig.~\ref{fig_synergy}. We draw an undirected edge between behaviors $i$ and $j$ iff $S_{ij}>\theta$, where we set the threshold $\theta=0.14$; edges below the threshold
are omitted to reduce clutter. Edge thickness is proportional to $S_{ij}$. 

\begin{table}[t!]
  \centering
  \footnotesize
  \setlength{\tabcolsep}{2.5pt}
  \begin{tabular}{lccccccc}
    \toprule
     & KVS & OVLP & HAE & MHO & CDO & KVCC & KVRM \\
    \midrule
    KVS  & -- & 0.09 & \cellcolor{mygreen}0.16 & 0.11 & 0.07 & 0 & \cellcolor{mygreen}0.14 \\
    OVLP & 0.09 & -- & \cellcolor{mygreen}0.42 & \cellcolor{mygreen}0.30 & \cellcolor{mygreen}0.38 & 0.06 & 0.05 \\
    HAE  & \cellcolor{mygreen}0.16 & \cellcolor{mygreen}0.42 & -- & 0.10 & \cellcolor{mygreen}0.53 & 0.02 & 0.06 \\
    MHO  & 0.11 & \cellcolor{mygreen}0.30 & 0.10 & -- & 0.10 & 0.06 & 0.10 \\
    CDO  & 0.07 & \cellcolor{mygreen}0.38 & \cellcolor{mygreen}0.53 & 0.10 & -- & 0.03 & 0.06 \\
    KVCC & 0 & 0.06 & 0.02 & 0.06 & 0.03 & -- & 0.02 \\
    KVRM & \cellcolor{mygreen}0.14 & 0.05 & 0.06 & 0.10 & 0.06 & 0.02 & -- \\
    \bottomrule
  \end{tabular}
  \caption{Normalized \rev{co-design affinity} matrix that encodes relative co-occurrence strength between behaviors across papers. Scores greater than the threshold $\theta = 0.14$ are \colorbox{mygreen}{highlighted} and visualized in Fig.~\ref{fig_synergy}.}
  \label{tab_norm_S}
\end{table}

\section{\rev{Extended Discussion on Challenges}}

Due to space constraints, this section complements \S~\ref{sec_challenge} with further discussion on open challenges.




\subsection{\rev{Trustworthy sKis}}
\label{a_trust}

As discussed in \textbf{C3}, efficiency optimizations typically account for average quality loss, but trustworthiness is rarely measured or attributed. Trust risks are especially concerning in high-stakes and safety-critical scenarios~\cite{wang-etal-2025-reasoning-enhanced, zeng2025futuresightdrive, wu2024safety, shen2026mftformer, li2025mmt, zhang2024yolo, sunmola2025surgical} where errors can have severe consequences. 

\rev{One representative example (also related to our discussion in \textbf{C3}) is that KV cache eviction and compression can compromise \emph{quality robustness}. They may drop rare but critical tokens with low accumulated attention (e.g., an exception clause in a contract, or a high value in a financial limit), which can lead to catastrophic errors on a small subset of inputs while the system still appears efficient and accurate on average. This failure mode can be amplified by distribution shift in workloads, such as in autonomous agent workloads~\cite{zhouwebarena, zhang2025ccma}, where statistically sparse tokens become logically important. Optimizations tuned to the original distribution may prune these sparse critical dependencies, causing agents to hallucinate success. Moreover, given that modern LLMs can be sensitive to prompt-level variations such as tone/politeness~\citep{cai2025does}, such structural methods may interact with these variations and affect robustness to prompt-level shifts.}

Trustworthiness risks extend beyond robustness to reliability, privacy, and safety, and can arise from diverse sKis behaviors. For instance, temporal asynchrony may expose stale KV and introduce nondeterminism, harming \emph{reliability}; cross-tier migration can leave residual KV state or transfer KV in plaintext, harming \emph{privacy}.

A key gap is that many methods only measure average metrics on relatively easy workloads, but rarely consider quality lower bound, recall SLO, or semantic violation metrics, so such worst-case failures remain invisible. A promising direction is to consider trustworthy metrics and integrate runtime mechanisms, such as violation detectors, recovery policies, and potentially targeted rectification mechanisms~\cite {yang2026attribution}, to provide a quality lower bound under stress.

\subsection{\rev{Intermediate Semantics for Behaviors}}
\label{a_intermediate}

We here provide additional discussion of intermediate semantics as a supplement to \textbf{C5}.

Intuitively, intermediate semantics for sKis behaviors aim to bridge the gap between binary decisions (e.g., ``retain'' vs. ``evict''); examples include ``reclaimable on GPU'', ``compressed on GPU'', ``compressed on CPU'', ``summarized on CPU/SSD'', etc. In this way, a co-optimization strategy can be formalized as a transition between these states. For instance, the compress-then-offload strategy first transitions a KV unit from a ``keep'' state to ``compressed on GPU'', then to ``compressed on CPU''. This creates a low-fidelity resident state that trades precision for I/O bandwidth. Similarly, lazy eviction transitions a KV unit to ``reclaimable on GPU'' with a grace period to allow cheap recovery before the final transition to permanent eviction (i.e., state ``evict''). These concrete examples show how future work can co-optimize eviction, compression, and migration by exploiting intermediate semantics.

\subsection{Benchmarking for sKis}

Here, we focus on system-performance benchmarking during serving, which measures actual performance metrics like latency, throughput, service-level objectives (SLOs), KV cache memory and bandwidth, and energy. In contrast, task or quality benchmarks focus on datasets and accuracy metrics. In our survey they serve only as quality gates and are not primary evaluation objectives. We refer readers to the survey~\cite{li2024survey} for details. 


\subsubsection{Review of sKis Benchmarking Practices}
\label{a_benchmark_review}

Many popular inference frameworks or systems, such as vLLM~\cite{vllm}, TensorRT-LLM~\cite{tensorrtllm}, and DeepSpeed-Inference~\cite{deepspeed}, provide benchmark scripts that measure system metrics for their local checks but remain framework-specific. We therefore view them as systems under test rather than the benchmark itself. 
In this section, we survey benchmarking efforts that provide a platform- and framework-agnostic way to obtain system measurements. They primarily fall into two categories: client-side tools and benchmark suites.

\paragraph{Client-side tools.} They define and enforce metric semantics. Using one tool across systems yields directly comparable numbers. 
LLMPerf~\cite{llmperf} targets API benchmark and provides system metric measurement on service endpoints.
NVIDIA NIM benchmarking guide~\cite{nvidia_benchmarking} defines the common metrics of time to first token (TTFT), end-to-end request latency, inter-token latency (ITL), tokens per second (TPS), and requests per second (RPS). The companion tool GenAI-Perf~\cite{genaiperf} emits the defined metrics and implements the stable-window analysis across OpenAI-API-compatible backends. However, although client-side tools offer specific metrics for LLM-based applications, we find inconsistent metric definitions and measurements across different tools.

\paragraph{Benchmark suites.} They are standardized packages of workloads, procedures, and reporting rules that specify what to run, how to run it, and what to report. Such suites typically cover multiple systems and hardware and enable reproducible comparisons.
MLPerf Inference~\cite{mlperf} emphasizes inference system comparison, and its v5.0 includes LLM scenarios with accuracy validation.
LLM-Inference-Bench~\cite{llminferencebench} evaluates the inference performance of the LLaMA model family across a variety of hardware platforms.
BALI~\cite{bali} measures LLM inference across six frameworks or acceleration approaches. It divides inference into three measured stages: setup, tokenize, and generate, and supports two settings: a technical setting with a fixed number of tokens, and a prompt-to-answer setting that includes tokenization.

\subsubsection{\rev{Benchmark Design Principles}}
\label{a_benchmark_future}

Building on the review, we offer a concise design checklist for sKis to improve comparability across systems. 
Here, we do not aim to define a complete, reproducible benchmark suite. Instead, we distill a set of benchmark design principles covering metrics, workloads, and reporting standards that future community benchmarks should instantiate.

\paragraph{Metrics.} Besides standard metrics, we recommend that future sKis benchmark suites report the following two types of important metrics:
\begin{enumerate}[leftmargin=*,itemsep=2pt,topsep=1pt,parsep=0pt]
    \item \textbf{Trustworthy metrics} that reflect the reliability of the serving system in satisfying SLOs, such as tail latency (P90, P95, P99 latency), SLO violation rate (\% of requests > P99 target), goodput (throughput meeting SLOs), recall SLO (success rate of certain semantic segments), and semantic violation rate.
    \item \textbf{KV-related resource metrics} that measure the utilization of resources, such as KV cache memory footprint (as \% of total GPU memory), average effective KV bitwidth (for compression methods), KV-related interconnect I/O (the volume of KV transferred across memory tiers), KV hit rate in memory tiers, KV-related stalls (\% of time spent waiting for KV transfers), effective bandwidth utilization (useful KV transfer ratio), and energy efficiency (Joules per token/request).
\end{enumerate}

\paragraph{Workloads.} A future sKis benchmark should cover the following three workload types to stress temporal, spatial, and structural KV behaviors:
\begin{enumerate}[leftmargin=*,itemsep=2pt,topsep=1pt,parsep=0pt]
    \item Multi-tenant or bursty online serving workloads to test the stability of temporal scheduling under high concurrency.
    \item Long-context task workloads to test KV cache placement and migration when memory and I/O become bottlenecks.
    \item Heterogeneous workloads, such as agent workloads~\cite{zeng2025janusvln, zhouwebarena}, structured editing~\cite{zeng-etal-2025-bridging}, or domain-shifted workloads~\citep{tian2024mokd, tian2026cross, li2023hong, li2025catch, yan2025hemora, zeng2025enhancing}, to test the robustness of structural KV cache optimizations against distribution shift.
\end{enumerate}

\paragraph{Reporting standards.} In addition to the basic information like model, hardware, and configuration, we recommend the following reporting standards for sKis benchmarks:
\begin{enumerate}[leftmargin=*,itemsep=2pt,topsep=1pt,parsep=0pt]
    \item Performance under graduated context lengths to validate scalability.
    \item Accuracy vs. memory curves for structural methods to reveal the trade-offs.
    \item Detailed hardware and topology setups, especially for temporal and spatial methods.
\end{enumerate}



\end{document}